\documentclass[10pt,twocolumn,letterpaper]{article}

\usepackage{cvpr}
\usepackage{times}
\usepackage{epsfig}
\usepackage{graphicx}
\usepackage{amsmath}
\usepackage{amssymb}
\usepackage{subcaption}
\usepackage{caption}
\usepackage{multirow}
\usepackage{authblk}

\usepackage[breaklinks=true,bookmarks=false]{hyperref}
\cvprfinalcopy 

\begin{document}

\title{Model Validation for Vision Systems via Graphics Simulation}

\author[1]{V S R Veeravasarapu\thanks{subbu@fias.uni-frankfurt.de}}
\author[1]{Rudra Narayan Hota\thanks{hota@fias.uni-frankfurt.de}}
\author[2]{Constantin Rothkopf \thanks{rothkopf@psychologie.tu-darmstadt.de}}
\author[1]{Ramesh Visvanathan \thanks{ramesh@fias.uni-frankfurt.de}}
\affil[1]{Center for Cognition and Computation, Goethe University Frankfurt}
\affil[2]{Institute of Psychology, Technical University Darmstadt}

\maketitle

\begin{abstract}
Rapid advances in computation, combined with latest advances in computer graphics simulations have facilitated the development of vision systems and training them in virtual environments. One major stumbling block is in certification of the designs and tuned parameters of these systems to work in real world. In this paper, we begin to explore the fundamental question: Which type of information transfer is more analogous to real world? Inspired from the performance characterization methodology outlined in the 90's, we note that insights derived from simulations can be qualitative or quantitative depending on the degree of the fidelity of models used in simulations and the nature of the questions posed by the experimenter. We adapt the methodology in the context of current graphics simulation tools for modeling data generation processes and, for systematic performance characterization and trade-off analysis for vision system design leading to qualitative and quantitative insights. In concrete, we examine invariance assumptions used in vision algorithms for video surveillance settings as a case study and assess the degree to which those invariance assumptions deviate as a function of contextual variables on both graphics simulations and in real data. As computer graphics rendering quality improves, we believe teasing apart the degree to which model assumptions are valid via systematic graphics simulation can be a significant aid to assisting more principled ways of approaching vision system design and performance modeling.

\end{abstract}

\section{Introduction}

Rapid advances in computation, combined with latest advances in computer graphics simulations have facilitated the development of vision systems and training them in virtual environments. 
Latest advances in computer graphics simulations has been leveraged for different phases of system design process, such as model learning, use in inference engines and validation etc. A recent interest in the deep learning community is shown to learn the models and parameters using 3D video games\footnote{futurism.com/videos/googles-deepmind-ai-has-now-moved-to-3d-games}.
However, one major stumbling block is in certification of the designs and tuned parameters of these systems to work in real world.
A fundamental open question is about the utility of graphics simulations for vision system design and validation.  The role of modeling and simulation in performance characterization of computer vision systems engineering was articulated in the early 90's  \cite{haralick1993performance,ramesh1995performance, christensen1997special}.  In these works, algorithms are viewed as statistical estimators and \textit{performance modeling} of an algorithm is viewed as a process of establishing the correspondence between the deviations of an algorithm's expected output as a function of the input signal parameters, perturbation model parameters, and algorithm tuning constants.  Simulation in this context was used as basis for demonstrating the robustness properties of the statistical estimator.   Our view is that in the system design process, computer simulations could allow a designer to perform a wide range of inferences, ranging from quantitative (for example, tuning/training the components in the context) to qualitative inferences (ranking/indexing the competing hypotheses or components for a task in the context). 
Recently, model-based graphics simulations are increasingly used in systems engineering and integrated into machine learning or probabilistic programming platforms \cite{kulkarnipicture} that allow numerical simulations to be tightly integrated into learning and inference. This opens up fundamental questions such as: \textit{What is the role of the current computer graphics tools for drawing conclusions for vision system design?  Are these conclusion drawn valid in reality? Which type of information transfer is more analogous to real world? etc}.

In this work, we address these questions with a perspective towards usage of graphics for vision system design, and our work is inspired from systems characterization standpoint. 
Our main focus is in exploration of utility of computer graphics tools for drawing conclusions in the system design process. In summary, our main contributions in the paper include: (1) adaptation of performance modeling methodology \cite{bb3334} to the context of using graphics to vision system design process. (2) Simulation based systems characterization gives varied degrees of insights from establishing correctness of implementation, to providing qualitative and quantitative insights. We demonstrate the utility of the graphics platform to provide qualitative to quantitative assessments of performance in a concrete and systematic validation setting, which considers multiple model hypotheses used in vision components design that are frequently found in generative model based vision literature. 
The study also compares qualitative and quantitative insights, obtained from simulated data, to insights from experiments on real world data, in order to assess the utility of physics-based graphics to vision validations. 

\begin{figure*}
\centering
\includegraphics[width=17cm, height=5cm]{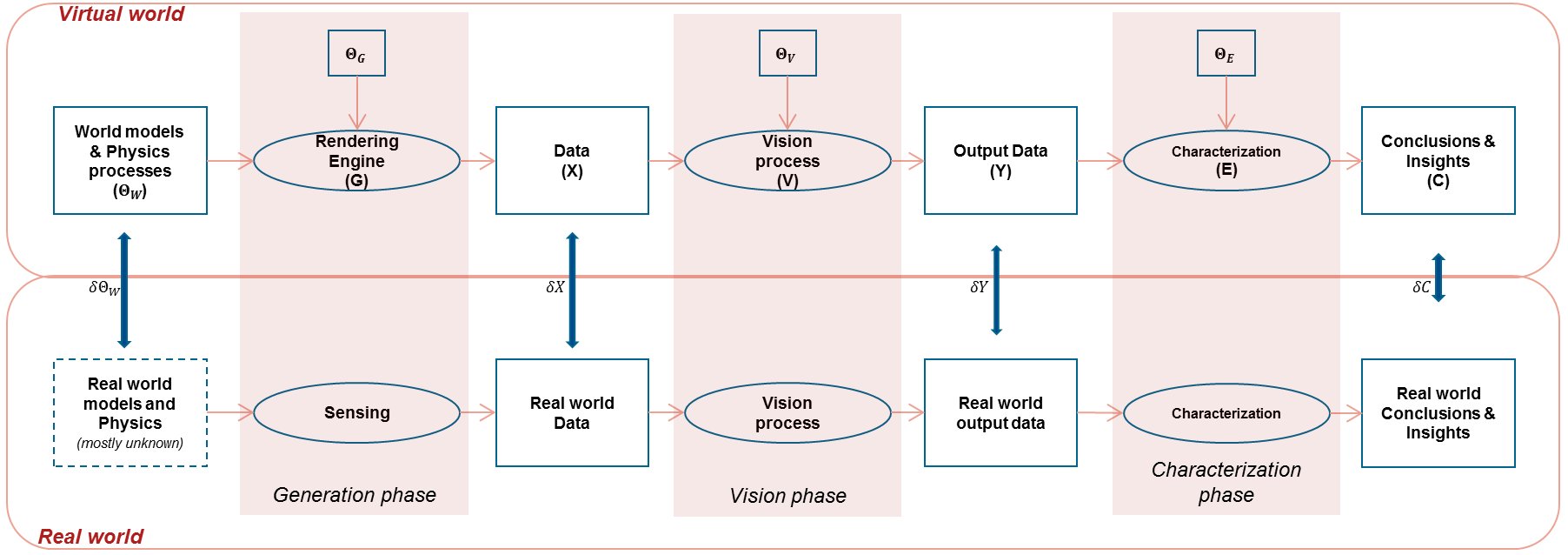}
\caption{\small Deviation (virtual vs real) propagation from graphics to vision: Processes (Generation, followed by Vision and Characterization) are shown with ellipses, while boxes represents input/outputs to the processes.}
\label{fig:virtual_vs_real}
\end{figure*}

\section{Performance Characterization Perspective}
A graphics simulation engine’s usefulness can be evaluated by thinking of it as a parametrized system whose input involving attributes such as: scene and object geometry, appearance, illumination, dynamics, environment, sensor and rendering parameters etc., are translated to image or video output. Deviations from reality in the inputs along with nature of computation used in the simulation engine map to deviations in the rendered output. These deviations in rendered data propagate through the subsequent stages to produce deviations in the final output. The significance of the impact of these deviations on the experimental conclusions depends on the nature of the conclusions an experimenter wishes to draw. These conclusions may range from qualitative to quantitative aspects of a vision system. 

We view \textit{using graphics for vision experimentations} as roughly a sequential process of information transfer with three steps as shown in Figure \ref{fig:virtual_vs_real}. 
(i) \textit{Generation-phase} is about specification of domain models of the application context in terms of deterministic or stochastic models of the scene parameters ($\Theta_W$) and rendering the scene to produce images/videos with required ground-truth. The deviations ($\delta \Theta_W$) in this virtual world models (from reality) propagate through rendering engine $G$, which is parametrized by $\Theta_G$, and produce deviations ($\delta X$) in the rendered data $X$. Minimizing these deviations at the image appearance level is one of the main interests of the computer graphics community (Photo-realism). 
(ii) \textit{Vision phase} depends on the experiment and task at hand. It can vary from validating model assumptions to systems of systems (end-user systems). This is denoted with $V$, might have some parameters $\Theta_V$. 
(iii) \textit{Characterization-phase}:- It is about scientific and rigorous analysis of task and context dependent performance and trade-offs for vision model or algorithm(s).  Thus, in order to characterize system behavior one specifies criterion functions $E$ (performance measures) that are to be measured through a carefully devised experimental protocol. More interesting empirical evaluations involve the estimation of how the criterion functions change as a function of design choices and tuning parameters (e.g. degree or rate of adaptation) of the system. Devising an experimental protocol for proper evaluation of an intelligent system is by itself a strong technical challenge. This phase might have some parameters $\Theta_E$ (for instance, thresholds of false-alarm and miss-detection rates etc.) which might be used to analyze trade-offs in the behavior of $E$ (as a function of $X$, $V$, $\Theta_V$) and draw some conclusions ($C$). 
The deviations in the rendered data propagate through the subsequent stages to produce deviations in the final conclusions $\delta C$. The magnitude of these deviations, now, depends on the invariance to fidelity in the steps, vision and inferences. For example, if vision process is using some spatial and/or temporal features which are invariant to appearance, then the deviations $\delta X$ will affect $\delta Y$ less. 
Hence, the significance of the impact of these deviations $\delta C$ on the experimental conclusion depends on task and the nature of the conclusions an experimenter wishes to draw.

More precisely, the deviations in experimental conclusions would vary depending on (a) closeness of the physics processes and models in the simulations to reality, (b) vision system algorithmic processes and their invariance to the other scene and graphics parameters and (c) type of conclusions (qualitative/quantitative) that one wants to draw. Again, from the view of performance characterization, these deviations can be formulated as a function ($F$) of model deviations, processes involved, and their free parameters.
\begin{equation}
\small{
\delta C = F (\delta \Theta_W, G, \Theta_G, V, \Theta_V, E, \Theta_E)
}
\end{equation}
This $F$ is highly complex and is a nonlinear function. Analytical derivation of $F$ might be difficult due to the complexity of light propagation in the scene, nonlinearities in rendering, vision algorithms and criterion functions used. Hence, we resort to empirical methods in this paper.

\subsection*{Generation phase}
The main purpose of computer graphics is to mimic (i.e. generate images of) 3D environments and data generative processes which dictates inputs and perturbations to the vision system. This phase consists of two steps: application domain modeling and rendering the images/videos, as shown in Figure \ref{fig:virtual_vs_real}.

\textbf{ Domain modeling} is about specifying stochastic/deterministic distributions of contextual variables ($\theta_W$) in the domain. It is the most critical part which directly influences the reliability of the conclusions. 
In this work, we start with modeling Manhattan worlds (object's coordinate systems are parallel each other and to world's coordinate system) using a sequential sampling procedure, that consists of (a) Road and street network generation, (b) Static object generation (c) Dynamic object and path generation, and (d) Other contextual parameter initialization and dynamics generation. Please refer to supplementary material for more elaborated discussions on domain modeling. 

\textbf{Rendering}:- A  physics based rendering platform has been discussed in the work \cite{Veeravasarapu:2015sv}, which facilitates sampling the contextual parameters ($\theta_W$) from the domain models ($p(\theta_W)$) to create 3D virtual worlds and renders the data along with required groundtruth. We use this platform to simulate the images/videos with required groundtruth depending the task at hand. The rendering parameters which influences the fidelity of simulated data (such as selection of graphics algorithms ($G$) and their parameters) are denoted as $\theta_G$. 
\begin{equation*}
\small{
\theta_W \backsim p(\theta_W)
}
\end{equation*}
\begin{equation}
\small{
\{X, O\} = G(\theta_W, \theta_G)
}
\end{equation}
where $\{X,O\}$ represents the set of inputs and corresponding ground-truth samples from graphics engine $G$. Please see the supplementary materials for the details of rendering processes used for this work.


\textit{Vision phase} facilitates the purposes of modeling, learning, inference, and validations in the system design process.  \textit{Characterization} is a process of establishing the performance of the model/module as a function of contextual parameters and its free parameters. The selection of performance or criterion measure ($E$) depends the task at hand and the insights or inferences that experimenter wishes to draw. For given contextual models and task, mathematically it can be formulated as,
\begin{equation}
\small{
E = f_V(\theta_W, \theta_V)
}
\end{equation}
where $f_V$ is a function in $n$-dimensional space ($n = |\theta_W|+|\theta_V|+|E|$, where $|.|$ denotes dimensionality of a vector) for a given model $V$ and a performance measure $E$. In this work, we consider scalar valued variables for the characterization experiments in the following sections. The function, $f_V$, can later be used for optimization purposes in design and/or parameter space. 



In the following section, we use this performance modeling methodology for a series of hypotheses validation experiments. To assess the utility of graphics for performance modeling in vision system design process, we compare the inferences derived from these experiments on simulated data to those made from real data.  
\begin{figure*}
\includegraphics[width=5.7cm,height=3.3cm]{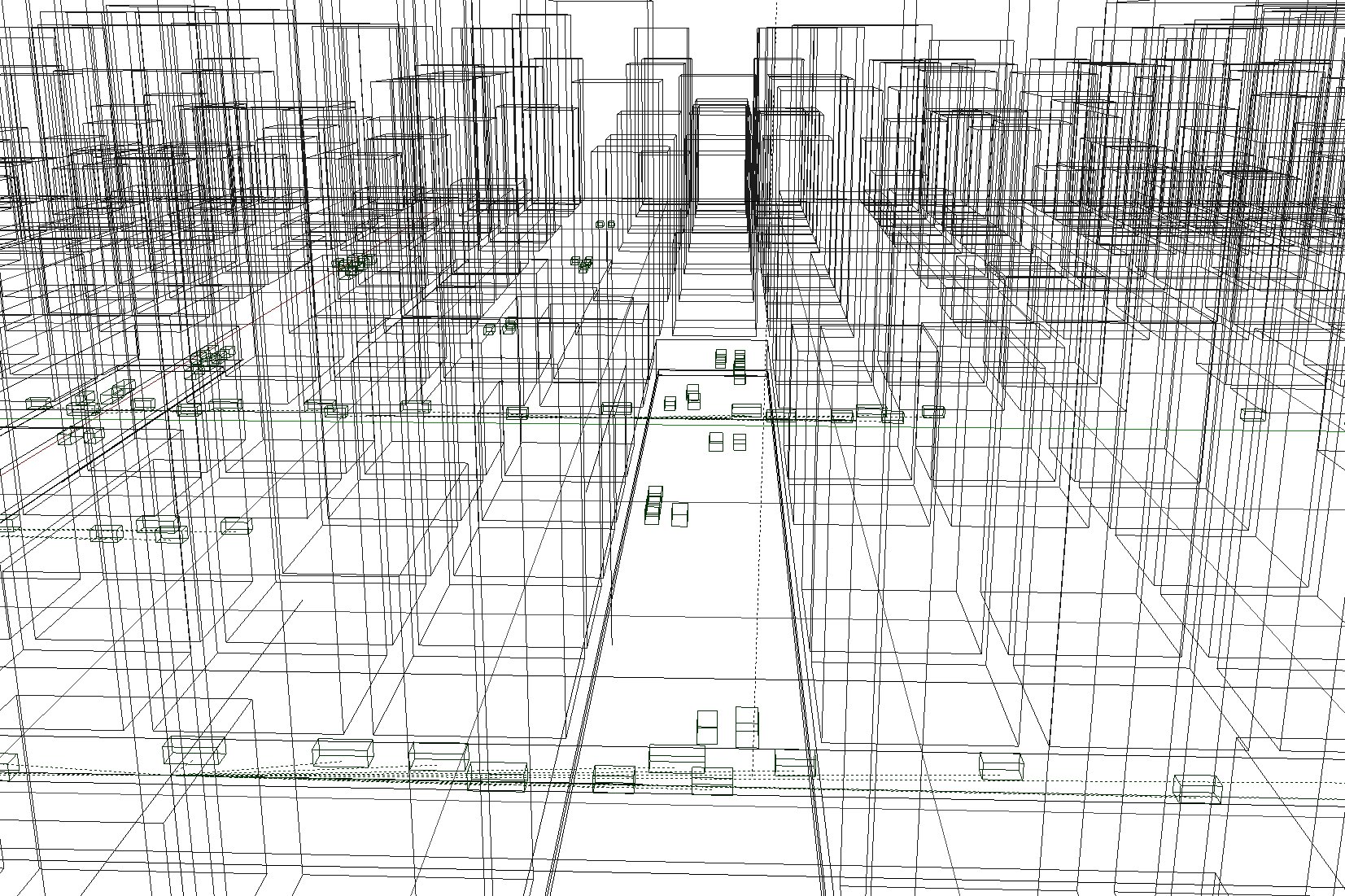}
\includegraphics[width=5.7cm,height=3.3cm]{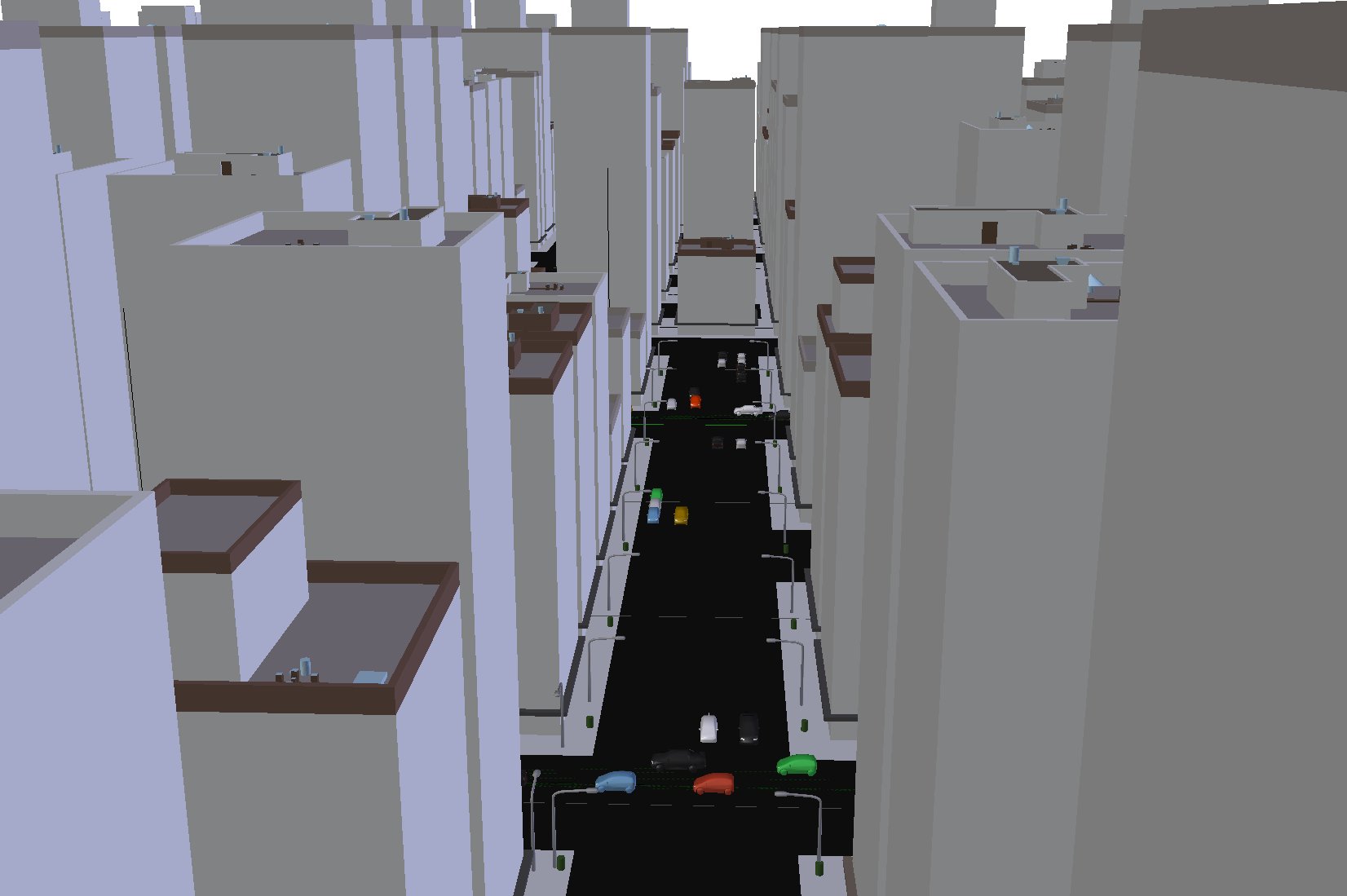}
\includegraphics[width=5.7cm,height=3.3cm]{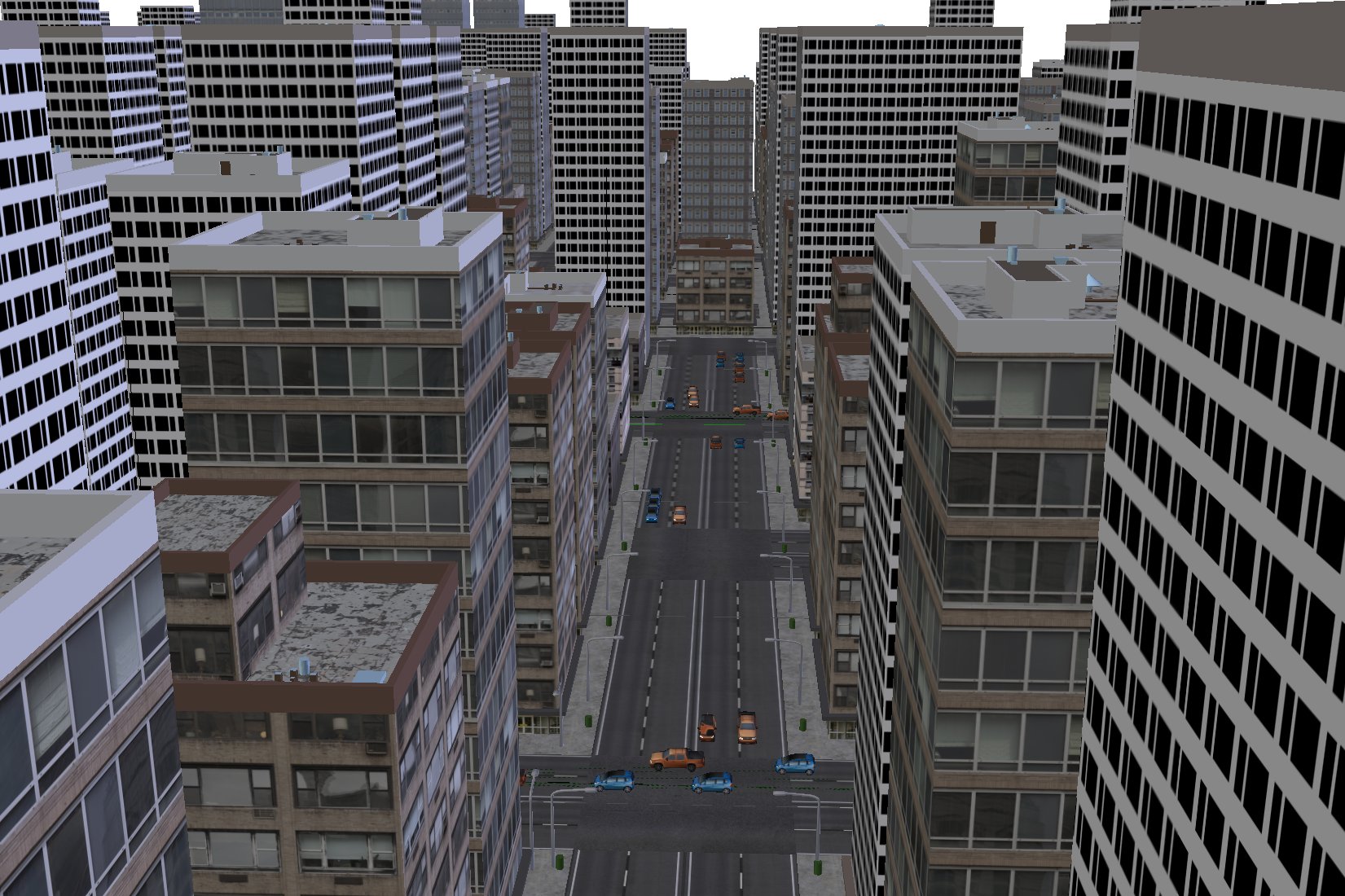}\\ 
\includegraphics[width=5.7cm,height=3.3cm]{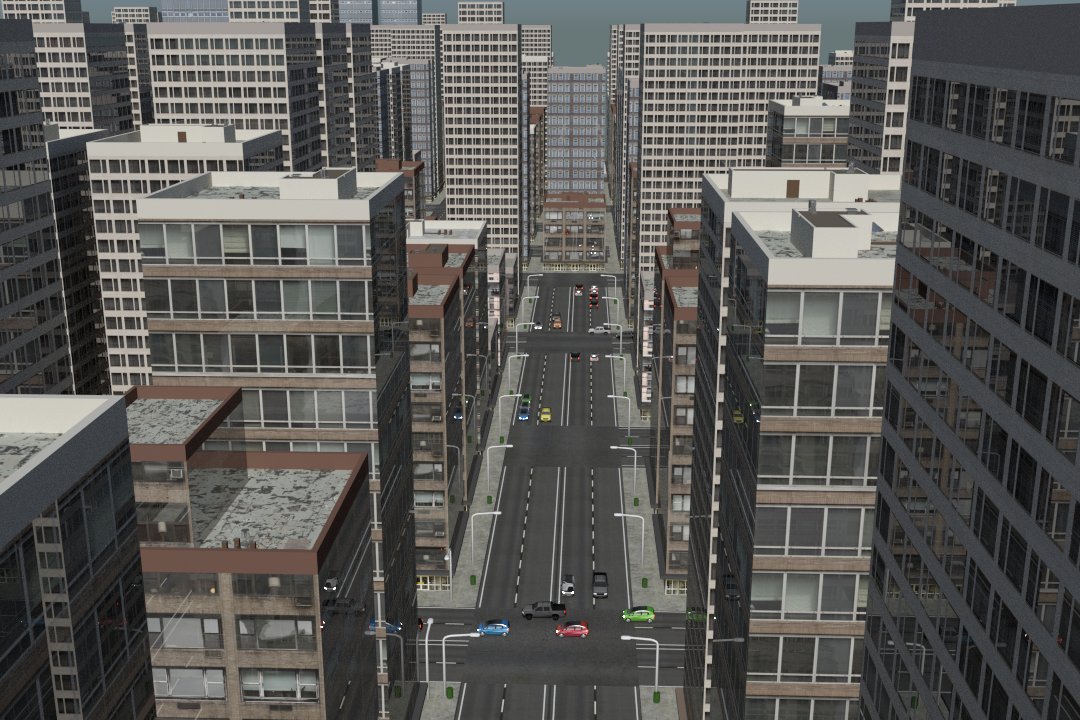}
\includegraphics[width=5.7cm,height=3.3cm]{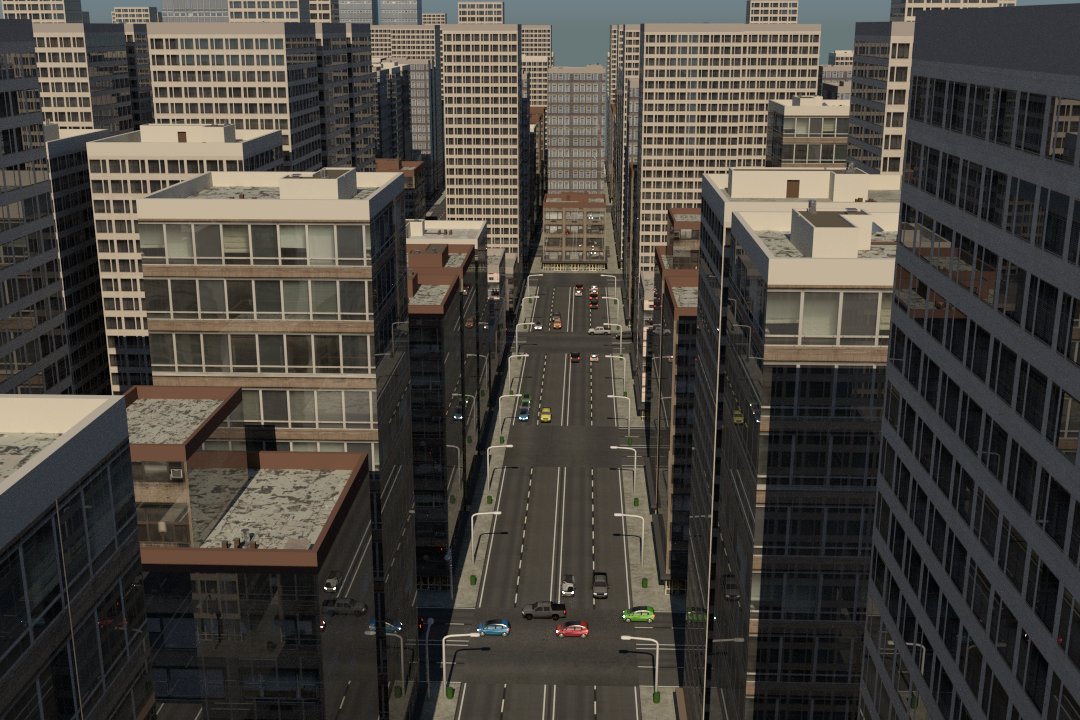}
\includegraphics[width=5.7cm,height=3.3cm]{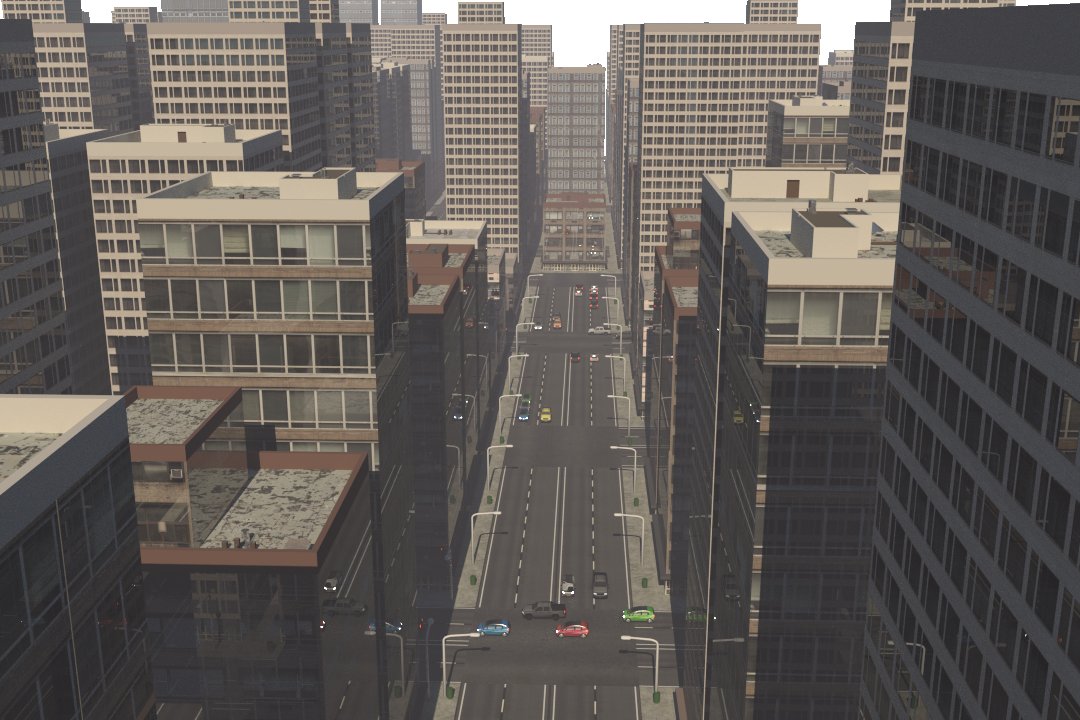}
\caption{\small{Manhattan city sample at multiple stages, \textit{Top row (left to right)}:- scene geometry generation (bonding boxes),  object surfaces, texture mapping; \textit{Bottom row (left to right)}:- Ambient illumination (overcast), sun-light and shadows, fog weather.}}
\label{fig_fidelity}
\end{figure*}

\section{Model Validation}
In this work, we use graphics simulations to establish the behavior of a multitude of models (selected from generative model based vision literature) as a function of its tunable parameters ($\theta_V$) and context ($\theta_W$). The main goal of these experiments are to find the types of contextual information and inferences that are richly analogous to real world. The set of hypotheses that we consider is provided in Table \ref{tab_hypothesis_set} along with related contextual causal variables, their model parameters and performance measures we use in the experiments.
\begin{table*}
\centering
\small
\begin{tabular}{|c|c|c|c|c|} 
\hline
\textbf{Model} &  Related contextual paras($\theta_W$) & Model parameters($\theta_V$) & Criterion measure (E) \\ \hline
Order Consistency Model (OC) & Photometry  & $s$ (Scale) & $\rho$ (Spearmann's rank correlation) \\ \hline
Brightness Constancy Model (BC) & Appearance  & $s$ (Scale) & $\sigma_{BC}^2$ (variance of pixel residual)\\  \hline
Gradient Constancy Model (GC) & Texture & $s$ (Scale) & $\sigma_{GC}^2$ (variance of gradient residual)\\ \hline
Piece-wise Smooth flow Model (PS) & Dynamics & $s$ (Scale) & $\sigma_{PS}^2$ (variance of flow residual)\\ \hline
Dichromatic Scattering Model (DS)&  Weather & - & $AE_{DS}$ (Average Angular Error) \\ \hline 
Object Shape Feature Model (OS) & Geometry $\&$  & $k$ (feature kernel) & $AMR_{OS}$ (Average Miss Rate)\\
 & Texture & HOG/LBP/HOG+LBP & \\ \hline 
\end{tabular}
\caption{\small{Hypotheses set: We select this set such that it covers hypotheses derived from different classes of physical contextual variables such as lighting, geometry (object shape), appearance, weather, and dynamics etc. Validation experiments of PS model is provided in the supplementary material.}}
\label{tab_hypothesis_set}
\end{table*}



\textbf{Order Consistency Model}:- 
Object detection in video surveillance systems is typically achieved through the use of background subtraction or change detection modules. The design of these modules involves modeling quasi-invariant measures which are insensitive to illumination changes and sensitive to geometric changes. A family of invariant operators \cite{singh2008order,xie2004sudden,bhat1998ordinal,zoghlami2009illumination} are derived from \textit{Order-consistency (OC) assumption}, e.g. the photometric transformations between image values of the same patch in 3D taken at two consecutive times is typically approximated as monotone.

\begin{figure*}
\begin{subfigure}[t]{1.0\textwidth}
\includegraphics[width=18cm,height=4.9cm]{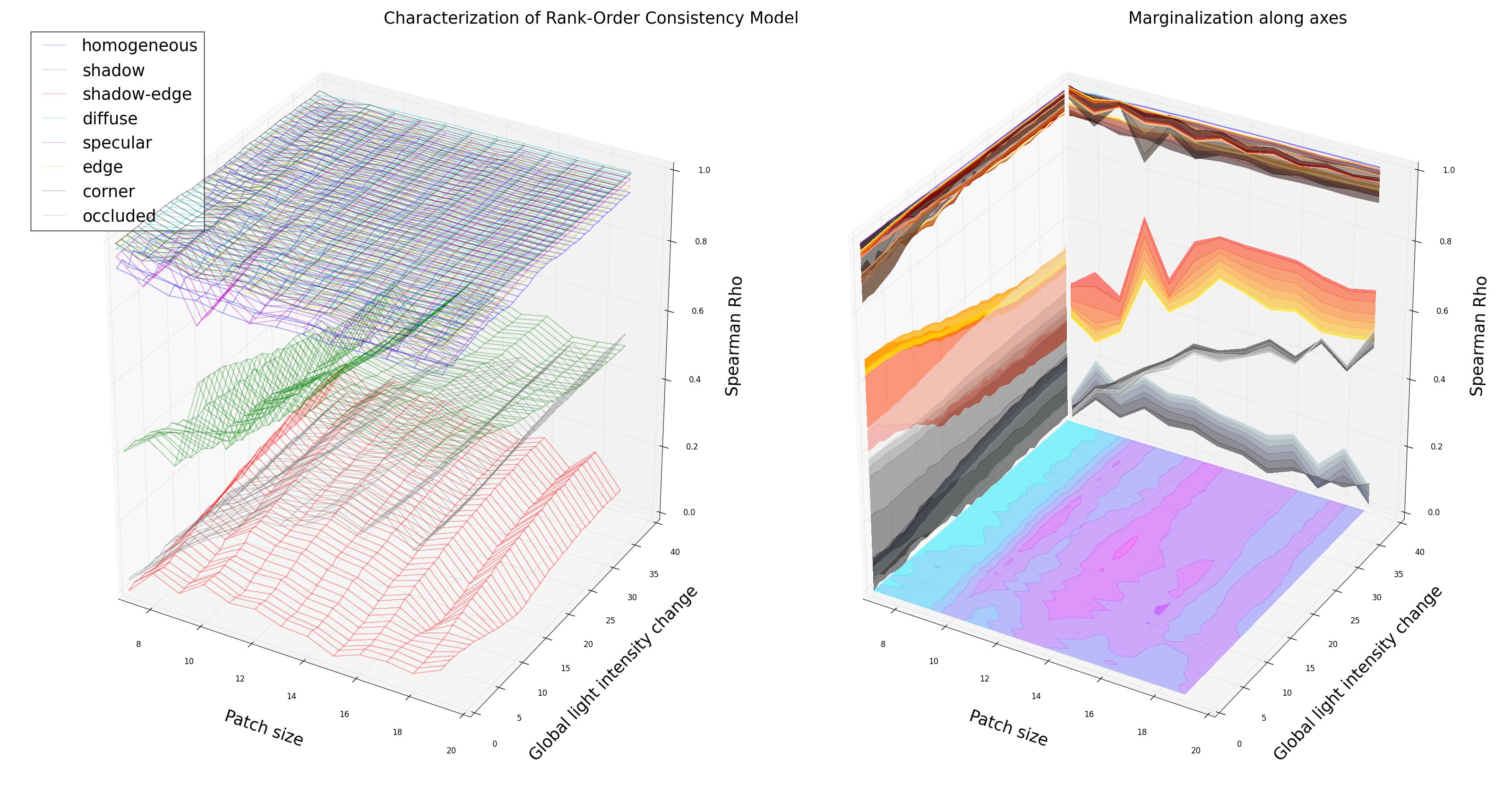}
\caption{Validation of OC model}
\label{fig_model_oc_global}
\end{subfigure}
\begin{subfigure}[t]{1.0\textwidth}
\includegraphics[width=18cm,height=4.9cm]{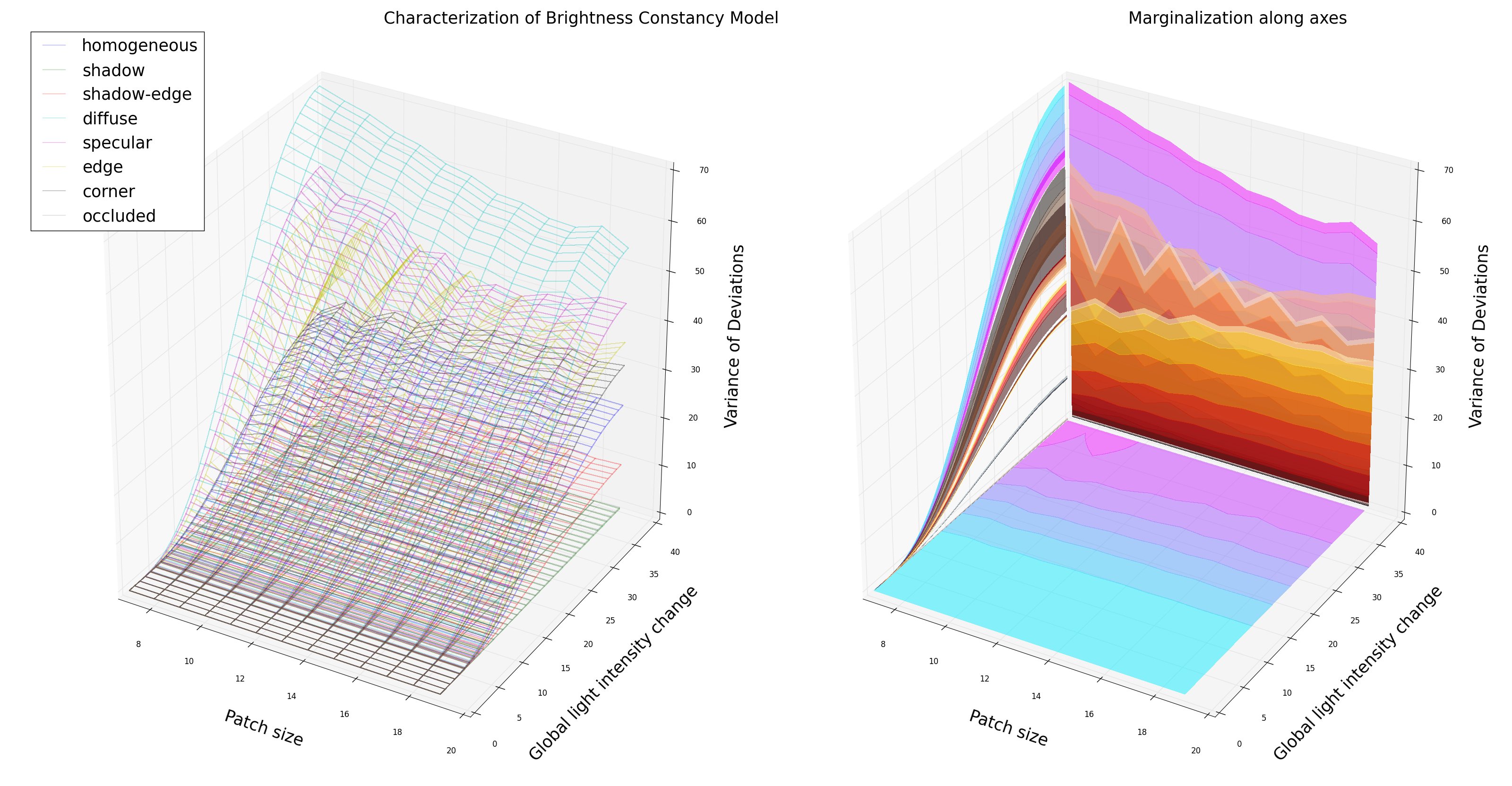}
\caption{Validation of BC model}
\label{fig_model_bc_global}
\end{subfigure}
\begin{subfigure}[t]{1.0\textwidth}
\includegraphics[width=18cm,height=4.9cm]{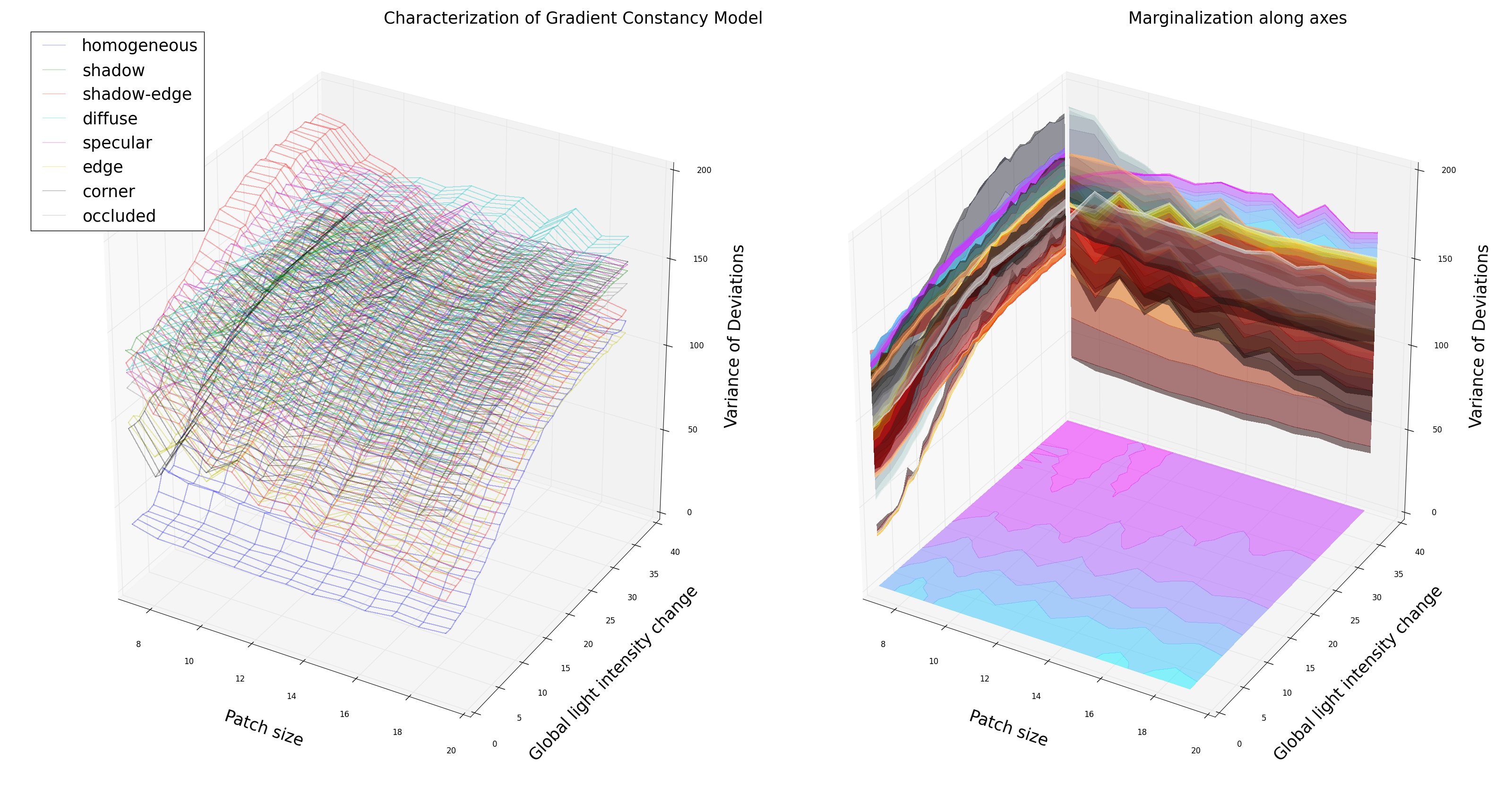}
\caption{Validation of GC model}
\label{fig_model_gc_global}
\end{subfigure}
\caption{\small Validations using simulations: Criterion measures are established as functions of contextual parameters and model parameters (left plots),  Right column plots are mere integration of characteristic manifolds of corresponding left plot, provided for better visual analysis of overall performance.} 
\label{fig_model_validations}
\end{figure*}

We validate this model under different contexts by establishing its behavior as a function of context and patch size. We choose the Spearman's rank correlation \cite{zar1972significance} coefficient ($\rho$) as a measure of monotonicity of the patch transformation. Other criterion measures can be considered depending on the task and user's interest. 
\begin{equation}
\small{
\rho = f_{OC}(\theta_W, s)
}
\end{equation}
where $\theta_W$ is a contextual variable that accounts for both spatial and temporal contexts. The nonparametric version of $f_{OC}$ is computed with simulations and shown in the Figure \ref{fig_model_oc_global}. The details of computation and inferences drawn from it are provided in the later sections.

\textbf{Brightness Constancy Model}: Motion detection in vision systems is typically achieved through the use of optical flow estimators or temporal differences. The design of these modules starts with \textit{Brightness constancy (BC) assumption}. Hence, the likelihood models might be derived from the term,  $\epsilon = I(i+u_{ij}, j+v_{ij}, t+1) -I(i,j,t)$. 
Its variance is used as a measure of deviation from the assumption and is characterized as a function of context and scale. 
\begin{equation} \label{eq_bc}
\small{ \sigma_{BC}^2 = f_{BC}(\theta_W, s) }
\end{equation}

\textbf{Gradient Constancy Model}:- Due to some decisive drawbacks of the BC model such as susceptibility to slight changes in brightness, the work \cite{uras1988computational} proposed to allow some small variations in the grey value and help to determine the displacement vector by a criterion that is invariant under grey value changes. Such a criterion is the gradient of the image grey value, which can also be assumed not to vary due to the displacement. The deviations from this gradient constancy (GC) assumption are measured as  $ \epsilon = \bigtriangledown I(i+u_{ij}, j+v_{ij}, t+1) - \bigtriangledown I(i,j,t) $. 
Hence, Its variance is used as a measure to its behavior and is characterized as a function of context and scale. 
\begin{equation} \label{eq_gc}
\small{
\sigma_{GC}^2 = f_{GC}(\theta_W, s) 
}
\end{equation}



\textbf{Dichromatic Scattering Model}:- 
Some works in the early 2000's, \cite{narasimhan2003models,schechner2001instant}, exploited physics based models for weather in order to estimate 3D scene properties such as depth. Most of these estimators are based on a color model for atmospheric scattering i.e, \textit{Dichromatic atmospheric scattering (DS) model}\cite{nayar1999vision}. It hypothesizes that the color of a given scene point ($\tilde{z}$) under bad weather is given by a linear combination of its actual color (after attenuation) , $z$ (as seen on a clear day), and airlight color, $z_a$ (color of fog or haze). Hence, $\overline{z}$, $\overline{z_a}$ and $\tilde{z}$ lie on the same plane in color space. This plane is called as \textit{Dichromatic plane}. Furthermore, the unit vectors $\overline{z}$ and $\overline{z}_a$, do not change due to different atmospheric conditions. Therefore, the colors of a scene point $x$, observed under different atmospheric conditions, are coplanar:
$\tilde{z} = m \overline{z} + n \overline{z}_a$,
here m and n are functions of scattering coefficient ($\beta$) of the weather and depth of the scene point. As a measure to the behavior of this model, we consider average angular error ($AE_{DS}$) of co-located pixels in five images of same scene (under different weather conditions) and the \textit{dichromatic plane} fitted using these 5 pixels  \cite{narasimhan2003models} in RGB color space.  Please note that the model parameters $m$, $n$ (functions of scattering coefficient and wavelength of the light) are covered by contextual parameters $\theta_W$. 

\begin{equation}
\small{
AE_{DS} = f_{DS}(\theta_W)
}
\end{equation}

\textbf{Object Shape Feature Model}:-
Learning object shape feature distribution (OS) models such as pedestrians and vehicles etc, is essential for object recognition  in surveillance and driver assistance applications. Learning shape feature based classifier models for pedestrian detection using virtual worlds is already explored by the works \cite{vazquez2014virtual,DBLP:journals/pami/XuRVL14}. They conclude that virtual and real-world based training give rise to similar classifiers. They considered three types of feature descriptors (k): (a) HOG, (b) cell-structured local binary patterns (LBP) and (c) combination of both (HOG+LBP). They used average miss-detection rate ($AMR_{OS}$) as a performance measure for testing shape model learnt in virtual worlds. We present their results and conclusions using our framework and, a more rigorous and systematic analysis is planned for the future. To harmonize with their experiments, we formulate the behavior of shape-feature model as
\begin{equation}
\small{
AMR_{OS} = f_{OS}(\theta_W, k)
}
\end{equation}
here $k$ is a feature descriptor. 

\subsection{Model Validations by Simulations }

\textbf{OC}: 
The similarity between order indexes of two patches can be measured with absolute spearman-rho ($\rho$). $\rho=1$ represents strict order-consistent, while $\rho=0$ represents perfectly non-consistent behavior. 
For the context of global illumination change, a series of images are rendered with increasing levels of light source intensity to mimic morning to noon sun variations. 
Reference image (scene without dynamic objects) is rendered under ambient illumination conditions.  
Patches from different spatial contexts (homogeneous region, shadow region, shadow boundaries, diffuse, specular surfaces, edge, corner and occlusions) are sampled in the images. For each spatial context, $\rho$ is computed between patch (sampled from image) and co-located patch in reference image. These values are plotted for different temporal contexts, in Figure \ref{fig_model_validations} (top-left plot). Data generation, rendering models and parameters used for simulations, and validations under other contexts are more elaborated in the supplementary. 

The left side plot of Figure \ref{fig_model_oc_global} shows the characteristic manifolds of the models for different spatial contexts under different levels of global illumination increments.  We see that the model is consistently performing better for lambertian (diffuse) surfaces (see the cyan-colored manifold in left side plot of Figure \ref{fig_model_oc_global}). The mean and standard deviation of $\rho$ values on diffuse surfaces are 0.98935 and 0.00423 respectively. We also experimented with a specular patch which is illuminated by reflections coming from vehicles (we treat vehicles as foreground objects). These patches behave just like occluded patches for which OC model fails (see gray-colored manifolds). The mean and standard deviation of $\rho$ for this case are  0.29344 and 0.13764 respectively. Observe the mean is too low and standard deviation is too high as the change is due to geometric transformation in the occluded patch. However, the model seems to be failing in the shadow regions. Shadow patch is nearly inconsistent (green-colored manifold) due to some sampling noise of rendering algorithm. This patch might improve its consistent behavior if we allow more rendering time. Shadow boundary patch behaves as order inconsistent (red-colored manifold) due to nonlinear illumination (direct and indirect illuminations). The mean and standard deviation of $\rho$ for the model under this context are  0.10022 and 0.05105 respectively. To analyze the overall and conditional performances of the model, we integrated the manifolds along the axes and projected on the corresponding planes (see right plot). For the projection on the ground plane, characteristic manifold of occlusion patch is skipped in integration, because it is with geometric change. The model is supposed to be consistent for photometric changes. The marginal variances of model deviations on diffuse, homogeneous patches are less, compared to occluded and shadow boundary patches. The model seems performing relatively better with patch size 13X13 for all levels of increments (magenta region on the ground plane). Hence, optimal patch size for this model is 13X13.  For lower patch sizes, the model is worse (cyan-colored region). 

To validate the utility of the graphics for global illumination change simulations for OC model validation, we compare these insights to that of similar real world sequences both quantitatively and qualitatively. We carefully selected the real videos from the benchmarking datasets \cite{goyette2012changedetection}, which capture similar temporal contexts considered above. Assuming first frame of these videos as reference image, we computed $\rho$ values, averaged over several patches sampled from each spatial contexts (similar to the ones considered in the above). These values are provided in Table \ref{tab_compare_oc} along with the ones on simulated data. From the table, it is clear that quantitative performance of the model is quite different across simulated and real world experimental settings. However, we observe that the qualitative statements and ordering of spatial contexts, made on the simulated data, are close to reality to some extent. 
More over, these insights also match with the statements found in the rank-order literature about its behavior on the type of patches \cite{singh2008order,xie2004sudden,bhat1998ordinal,zoghlami2009illumination,mittal2006intensity}.

\begin{table}
\centering
\small
\begin{tabular}{|p{1.7cm}|p{2cm}|c|c|} \hline
Spatial context & Temporal context & $\rho$ (simulated) & $\rho$ (real) \\ \hline \hline
Homogeneous & Global Illumination change & 0.7868 (4) & 0.4457 (5) \\ \hline
Diffuse & Global Illumination change & 0.8323 (2) & 0.5968 (4)\\ \hline
Shadow boundary & Global Illumination change & 0.0877 (6)& 0.6046 (3)\\ \hline
Edge & Global Illumination change & 0.8076 (3)& 0.8313 (1)\\ \hline
Corner & Global Illumination change & 0.8350 (1)& 0.7574 (2)\\ \hline
Occluded  & Global Illumination change & 0.2622 (5) & 0.2635 (6) \\ \hline 
\hline
Over all & Day light & 0.6691 (1) & 0.6472 (1) \\ \hline
Over all & Night & 0.2386 (3)& 0.2550 (3)\\ \hline
Over all & Fog & 0.4618 (2)&  0.5429 (2)\\ \hline 

\end{tabular}
\caption{\small Comparison of average of $\rho$ values for different contexts across real and simulated sequences. Numbers in the brackets are ranks.}
\label{tab_compare_oc}
\end{table}


\textbf{BC}: 
To validate the BC model in the context of global illumination change, we consider a scene with moving vehicles from the Manhattan city model. 40 versions of second frame is rendered under varying light intensity levels as mentioned above.  To validate the BC model in different spatial contexts, we compute the variance of the residuals ($I_t(i,j) - I_{t+1}(i+u_{ij}, j+v_{ij})$) between gray values of the pixels of the first patch and same pixels (moved by flow vector) in the second frame (see Eq.\ref{eq_bc}). In the left plot of the Figure \ref{fig_model_bc_global}, variance ($\sigma^2_{BC}$) is plotted against global light intensity levels and patch size for different types of patches (just as mentioned above) and right plot shows marginalizations along the axes. From these plots, BC model shows consistent behavior for very few levels of increment in the light source intensity levels for all spatial contexts and fails for large increments which is quite obvious. Interestingly, the model is quite robust for shadow regions due to indirect illumination. The model seems to be failing for all other spatial contexts that are illuminated directly by the light source. Surprisingly, diffuse patch is having more deviations for this temporal behavior of the source. As the model is based on pixel level statistic (gray-value), patch size has insignificant effect on the behavior of the model. 

\textbf{GC}: To validate the GC model in different spatial contexts, we compute the variance of the residuals ($\bigtriangledown I_t(i,j) - \bigtriangledown I_{t+1}(i+u_{ij}, j+v_{ij})$) between intensity gradient values of the first patch and same pixels (moved by flow vector) in the second frame (see Eq.\ref{eq_gc}). Second frame is simulated under gradually varying sun illumination as above mentioned. In the left plot of the Figure \ref{fig_model_gc_global}, variance ($\sigma^2_{GC}$) is plotted against global light intensity levels and patch size for different types of patches (just as mentioned above) and right plot shows marginalizations along the axes. This model seems to be quite stable compared to BC model for all spatial contexts . Good performance of the model is found on the homogeneous regions (blue manifold in the plot) which is obvious due to less gradient values. Patches with shadow edges, building surface edges and occlusions are having more deviations  in the model. 



\textbf{DS}: For the validation of dichromatic scattering (DS) model, we use five images each rendered under different foggy, misty, rainy and hazy conditions using MC path tracer with 200 render samples. Physics-inspired weather models used for simulations are explained in the supplementary. Like in \cite{narasimhan2003models}, the dichromatic plane for each pixel was computed by fitting a plane to the colors of that pixel, observed under the five atmospheric conditions. The error of the plane-fit was computed in terms of the angle between the observed color vectors and the estimated plane. The average angle error (AE in degrees) for all the pixels in each of the five images is shown in Table \ref{tab_model_ds}. 
On simulated data, model seems to be working better for fog, mist, rain and dense haze under ambient light. For mild haze conditions under a point light source (sun), the model does not perform well. 
In their work, Narasimhan et al, \cite{narasimhan2003models}, already tried to validate the model for different weather conditions using multiple real world images taken under each weather condition. Experiments were performed using the scene imaged 5 times under each of the different foggy, misty, rainy and hazy conditions. These average angular error values are taken from \cite{narasimhan2003models} and provided in Table.\ref{tab_model_ds} for comparing simulation results with that of real world.  The quantitative error metrics are different for the both worlds, most likely due to mismatch of contextual models and rendering approximations. We also tried to rank the different contexts w.r.t. correctness of the model, and this ranking seems to be similar across synthetic and real data. 

\textbf{OS}: Pedestrian detectors are trained on the simulated data ($V_{tr}$), generated with \textit{Half-Life2} video game by city driving \cite{vazquez2014virtual}. To cope up with the experiments on real world setting, the size of training data-set is limited to 2416 pedestrian images and 1218 pedestrian-free images of $640X480$ resolution. Two real world benchmark data-sets are used for the comparison: INRIA ($I$) and Daimler ($D$). They are divided into training ($I_{tr}$, $D_{tr}$) and testing ($I_{tt}$, $D_{tt}$) sets. In order to detect pedestrians, given test image is scanned for obtaining windows to be classified as containing a pedestrian or not by a classifier learnt on training data. Single best window is selected, since multiple detections can be due to a single pedestrian. Please see\cite{vazquez2014virtual} for the details of employed scanning and selection procedures, and experimental settings and real world data-sets etc. 

Table \ref{tab_pedestrian_model} shows the results of three classifiers in different training and testing settings. Each data-set has different contextual models underlying their acquisition processes. we assume that these data-sets are three realizations of different contexts. These differences in contextual models result in dataset-bias \cite{torralba2011unbiased} problem, not only in between virtual and real world training settings but also in between real world and real world training settings. First two rows in the table show the results of the classifiers on the real world data when they trained on virtual data. These quantitative figures are similar to the results when they trained and tested on real world data but from different data-sets.  

On the basis of dataset-bias, the virtual-world domain is comparable to a real-world one for learning shape feature models. Three classifiers are ranked according to AMR values. These rankings seem to be similar across the domains. HOG+LBP is performing better consistently. HOG is poor but giving average performance when training and testing data-sets matched. The rankings of these feature descriptors are matching when tested on INRIA dataset after training in virtual and real (INRIA) world.  Same behavior is observed when transferring from virtual to Daimler dataset. Validation experiments of PS model and More rigorous analysis of characteristic manifolds are presented in the supplementary material.

\begin{table}
\footnotesize
\centering
\begin{tabular}{|p{1.2cm}|p{1.2cm}|p{1.1cm}|p{0.4cm}|p{0.4cm}|p{0.7cm}|p{0.4cm}|}
 \hline
\multicolumn{3}{|c|}{\textbf{Context}} & \multicolumn{4}{|c|}{\textbf{Model Errors}} \\ \hline
 & & & \multicolumn{2}{|c|}{\textbf{Real}} & \multicolumn{2}{|c|}{\textbf{Virtual}} \\ \hline
weather & particle radius($\mu m$) & density ($cm^3$) & AE & E$<3^o$ & AE & E$<3^o$   \\ \hline \hline
Fog (Ambient) & 1-10 & 100-10  & $0.58^o$ (1)& $95\%$ (1) & $0.1373^o$ (1) & $100\%$ (1) \\ \hline
Mist (Ambient)& 0.1-1 & 100-10  & $1.25^o$ (3)& $88\% $ (3) &$0.3887^o$ (2)&$97\%$  (2)   \\ \hline
Rain (Ambient)& $10^2$-$10^4$ & $10^{-2}$-$10^{-5}$  & $1.13^o$ (2)& $91\%$ (2)& $1.2434^o$ (4)&$94\%$ (4)   \\ \hline
Dense Haze (Ambient) & $10^{-2}$-$1$  & $10^3$-$10^2$ & $2.27^o$ (4) &$76\%$ (4)& $1.0122^o$ (3) & $95\%$ (3) \\ \hline
Mild Haze (Sunny)  & $10^{-2}$-$1$ & $10^2$-$10$ & $3.61^o$ (5)& $44\%$ (5) & $2.4563^o$ (5)& $78\%$ (5)   \\ \hline
\end{tabular}
\caption{\small Validation of the dichromatic model with the scene rendered 5 times under each of the different foggy, misty, rainy and hazy conditions. 
}
\label{tab_model_ds}
\end{table}

\subsection{Domain Adaptation}
In the above experiments, most of qualitative statements about models and their indexing results (ranking contextual models w.r.t. vision model or vice-versa) match with the reality. However, quantitative analytics (performance measures) are quite biased. Simulated images, although photo-realistic, come from a different generation system than those acquired with a real camera. Deviations in contextual models, approximations in rendering and different noise sources in graphics pipeline may end up as a dataset-bias problem, which in turn results in some bias in conclusions from simulations. The bias due to mismatch between contextual models is quite common even in between any two real world datasets. The actual bias due to approximations in rendering engine is what we interested in. To measure this bias (only due to the graphics), a careful systematic settings is needed, where virtual world models can be learnt from the real world data.  We plan to explore this in detail in future work. However, this bias might be corrected to some extent by using the concepts from domain adaptation, i.e, adding a few real world samples to synthetic data. For example, see the last two rows of Table \ref{tab_pedestrian_model}, containing the results for which  $10\%$ of the training data is taken from the real data ($I_{tr}$ or $D_{tr}$) and added to the virtual data. This addition of INRIA data turns out in improvements by 10, 6.04 and 9.11 points for three classifiers respectively. Similar behavior is observed also for Daimler data. Hence, the work \cite{vazquez2014virtual} concludes that the simulations allow to significantly save the costs for real world data acquisition and annotations efforts.

\begin{table}
\footnotesize
\centering
\begin{tabular}{|p{2cm}|c|c|c|} 
\hline
Context$\downarrow\ \ \ $ & HOG & LBP & HOG+LBP \\ \hline
$V_{tr} \Rightarrow I_{tt}$ & $32.47 \pm 0.47 $ & $28.87 \pm 0.70$ & $23.81 \pm 0.53$ \\  & (3) & (2) & (1) \\ \hline
$V_{tr} \Rightarrow D_{tt}$ & $30.64 \pm 0.43 $ & $45.21 \pm 0.49$ & $28.27 \pm 0.48$ \\  & (2) & (3) & (1) \\ \hline \hline
$I_{tr} \Rightarrow I_{tt}$ & $21.27 \pm 0.52 $ & $18.42 \pm 0.53$ & $14.35 \pm 0.46$ \\ & (3) & (2) & (1) \\ \hline
$D_{tr} \Rightarrow D_{tt}$ & $30.01 \pm 0.51 $ & $35.07 \pm 0.29$ & $22.48 \pm 0.45$ \\ & (2) & (3) & (1) \\ \hline \hline

$I_{tr} \Rightarrow D_{tt}$ & $41.12 \pm 1.01 $ & $35.40 \pm 0.70$ & $26.22 \pm 0.85$ \\& (3) & (2) & (1) \\ \hline
$D_{tr} \Rightarrow I_{tt}$ & $38.46 \pm 0.45 $ & $39.54 \pm 0.55$ & $32.28 \pm 0.47$ \\ & (2) & (3) & (1) \\ \hline \hline

$(V_{tr},10\% I_{tr})$ & $22.47 \pm 1.01 $ & $22.83 \pm 0.92$ & $14.70 \pm 0.63$ \\ $\ \ \ \ \ \ \ \ \ \ \Rightarrow I_{tt}$ & (2) & (3) & (1) \\ \hline
$(V_{tr},10\% D_{tr}) $ & $26.13 \pm 0.66 $ & $34.69 \pm 1.15$ & $21.71 \pm 0.73$ \\$\ \ \ \ \ \ \ \ \ \ \Rightarrow D_{tt}$& (2) & (3) & (1) \\ \hline
\end{tabular}
\caption{\small Training ($tr$) and Testing ($tt$) of pedestrian shape models:: $V$, $I$ and $D$ represent Virtual, INRIA \cite{dalal2005histograms} and Daimler \cite{enzweiler2009monocular} pedestrian data-sets respectively.  }
\label{tab_pedestrian_model}
\end{table}

\section{Conclusions}
In this work, we try to begin to address the question, \textit{Which type of information transfer from graphics is more analogous to real world?}. 
The main focus of this paper is to examine invariance assumptions used in video surveillance settings as a case study and evaluate the degree to which those modeling assumptions hold in graphics simulations and in real data. 
From the results of our experiments, we observe that the effect of fidelity of the graphics (photorealism) on the performance of vision system (in real world) is dependent on the type of information transfer from graphics to vision and closeness of distributions of models used for simulations to reality. In our specific context of experiments, vision models are meeting qualitative expectations on the overall performance, but relation between spatial contexts and model behavior is quite affected in the virtual worlds for the models like OC, BC, GC, and DS models. Qualitative inferences such as ranking the models like object shape feature models in the given context are quite similar across real and virtual worlds. However the deviations and bias in the conclusions about these kind of appearance-quasi-invariant models can be corrected to some extent using domain adaptation concepts. 
Future research is needed to tease apart the degree of invariance as a function of time and contexts (e.g. shadows, reflections, weather state, dynamic range of camera irradiance, etc.). We believe that teasing apart the degree to which model assumptions are valid via systematic graphics simulation can be a significant aid to assisting more principled ways of approaching vision system design and performance modeling.

\section*{Acknowledgments}
This work was supported by the German Federal Ministry of Education and Research (BMBF) in
the projects, 01GQ0840 and 01GQ0841 (BFNT Frankfurt).

{\small
\bibliographystyle{ieee}
\bibliography{egbib}
}

\appendix
\section*{Appendix}

The appendix is organized as follows. In Section \ref{sec_domain_modeling}, we provide the details of the application domain models, where as Section \ref{sec_rendering} gives the details of rendering processes and their parameters, that are used to generate the data for the experiments, discussed in the main paper. Section \ref{sec_ps} validates piece-wise smooth flow (PS) model on simulated samples.  

\section{Application Domain Modeling} \label{sec_domain_modeling}
Application domain modeling is about specifying stochastic/deterministic
distributions of contextual variables ($\theta_W$) that are specific to the application domain and task. In this work, we discuss domain modeling specific to (Manhattan) outdoor surveillance applications. Modeling dynamic environments is an essential task for various learning, validation applications of vision systems and also for virtual game applications and urban planning.  Creating these environments requires a lot of man hours from many designers. Recently, vision researchers started using existing open-source or commercial virtual game engines as application domain simulators, see for instance, \textit{ObjectVideo} \cite{taylor2007ovvv} and \textit{VDrift} \cite{haltakov2013framework,espie2005torcs} etc. These simulators are proven to be helpful to characterize contextual domain priors. However, the games are built with fake appearance (mathematically simplified) methods \cite{pharr2004physically} and heuristics for efficient rendering that are not physically correct.

For a structured design, we divide the physical contextual variables into categories such as: (a) \textit{Geometry} (scene structure and object shapes),  (b) \textit{Photometry} (includes extrinsic and intrinsic parameters of light sources, sensors, weather states in the scene and materials of the objects), (c) \textit{Dynamics} (Temporal variations of above parameters). 
 These can be specified in several ways: Manual scripting using probabilistic programming platforms \cite{mansinghka2013approximate}, learning from real world data \cite{hahnel2003learning}, adopting from games \cite{pelechano2005crowd}, or in a mixed manner (for example, the partial information such as weather dynamics and illumination priors could be learned from real world data while other information such as geometry can be manually specified). (Semi) automatic large content creation with procedural grammars has been attempted in \cite{parish2001procedural} and a recent survey of these methods can be found in \cite{smelik2014survey}. However, they might produce highly correlated scenes for two runs of the systems as scene generation is based on given axioms and deterministic logic. Another way is to use point processes to generate spatial and temporal patterns of a dynamic scene environments. The parameters of these point processes can be learnt from the real data and set by experts.

 Modeling 3D domains is getting easier due to the availability of large-scale 3D object repositories such as Google's 3D warehouses. In this work, we model Manhattan world geometries (object's coordinate systems are parallel each other and to world's coordinate system). We use marked point processes to generate stochastic Manhattan city geometries, followed by importing existing 3D objects into the sampled scene geometry (3D bounding boxes).

\subsection{Geometry}
\subsubsection{Scene geometry with Marked Point Processes}
A prior knowledge about spatial/temporal distributions of a set of objects can be modelled with stochastic processes such as point processes \cite{jacobsen2006point}. A realization of the point process consists of a random countable set of points $\{ o_1,...,o_n \}$ in a bounded region $\textbf{O} \in R^d$. A marked point process (MPP) couples a spatial point process $Z$ with a second process defined
over a \textit{mark} space $M$ such that a random
mark $m_o \in M$ is associated with each point $o \in Z$. For example, a 3D point process of cuboid marks has elements of the form $o_i = (p_i, ( c_i, l_i, b_i, h_i, \theta_i))$ specifying the
location ($p$), object class ($c$), dimensions ($l$,$b$,$h$) and orientation ($\theta$) of a specific bounding box (cuboid) in the scene. In this work, we use MPP to incorporate our prior knowledge about the spatial patterns in Manhattan city geometries (extrinsic size and transformation of objects such as buildings and vehicles etc). Thus, the realization of the MPP in this work consists of an object location $p$ defined on a
bounded subset of $R^2$ ($xz$-plane), together with a mark $m$ defining type and a
geometric 3D shape to place at point $p$. In other words, we model the objects in the scene as a set of configurations from an MPP that incorporates prior knowledge such as expected sizes of buildings, trees, people and vehicles on the scene of knowledge about the scene regions where these objects will not appear. We denote the prior term for an object as $\pi(o_i)$, and assume independence among the objects. The mark process is assumed as independent from the spatial point process, so that priors in MPP would be factored as: $\pi(o_i) = \pi(p_i)\pi(m_i)$. However, this common approach ignores obvious and strong correlations between the size and
orientation of objects. Hence, a conditional mark process is introduced for cuboids representing shape and orientations of a 3D bounding box, conditioned on the location and shape, leading to a factored prior of the form:

\begin{equation}
\small 
\pi(o_i) = \pi(p_i, h_i, \theta_i | c_i, l_i, b_i)  \pi(l_i|c_i) \pi(b_i|c_i) \pi(c_i) 
\end{equation}
The prior for object class (type such as building, tree, pedestrian and vehicle classes etc) distributions are modelled with uniform multinomial distribution. This means object type follows a uniform distribution, and given object type, the shape dimensions (3D bounding box) are \textit{i.i.d}. One can also learn the parameters of these priors through passive training using real world data. 

\textbf{The conditional mark process}: We represent prior for $\pi(p_i, h_i, \theta_i | c_i, l_i, b_i)$ as independent uniform distributions on the bounded regions. $\theta_i$ is fixed for this work to model Manhatten city in which object's coordinate system is parallel to the world's coordinate system. The nonoverlap between any two objects is incorporated by using a lookup map of the bounded region (region occupancy map). The parameters of 3D bounding boxes given the object class are modeled by simple Gaussian distributions. 

\begin{figure*}
\centering
\begin{subfigure}[t]{1.0\textwidth}
\includegraphics[width=17cm, height=5cm]{images/fidelity/bounding_box.jpg}
\caption{\small 3D bounding boxes sampled from Marked Point Process}
\end{subfigure}
\begin{subfigure}[t]{1.0\textwidth}
\includegraphics[width=17cm, height=5cm]{images/fidelity/textures.jpg}
\caption{\small 3D object models are resized and fitted in bounding boxes}
\end{subfigure}
\begin{subfigure}[t]{1.0\textwidth}
\includegraphics[width=17cm, height=5cm]{images/fidelity/ambient_rgb_while.jpg}
\caption{\small rendered under ambient light}
\label{fig_render_ambient}
\end{subfigure} 
\begin{subfigure}[t]{0.32\textwidth}
\includegraphics[width=5.5cm, height=4cm]{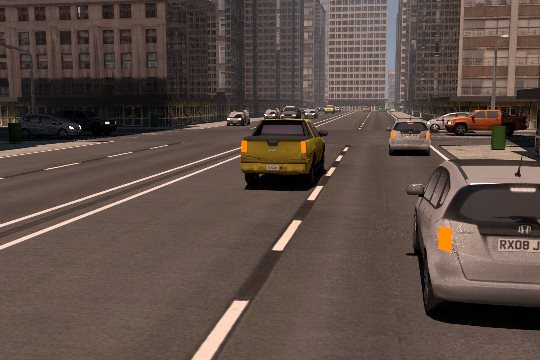}
\caption{\small a sample from rendered data for automotive applications}
\label{fig_auto}
\end{subfigure}
\hspace{1mm}
\begin{subfigure}[t]{0.32\textwidth}
\includegraphics[width=5.5cm, height=4cm]{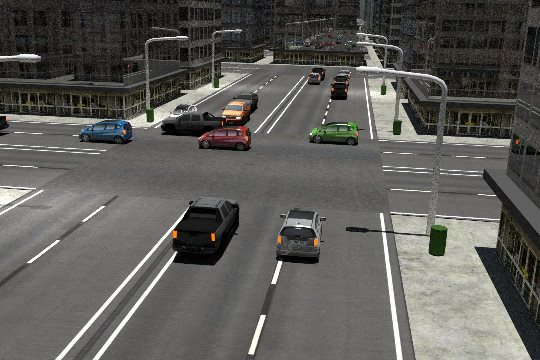}
\caption{\small a sample from static surveillance data}
\label{fig_surveillance}
\end{subfigure}
\hspace{1mm}
\begin{subfigure}[t]{0.32\textwidth}
\includegraphics[width=5.5cm, height=4cm]{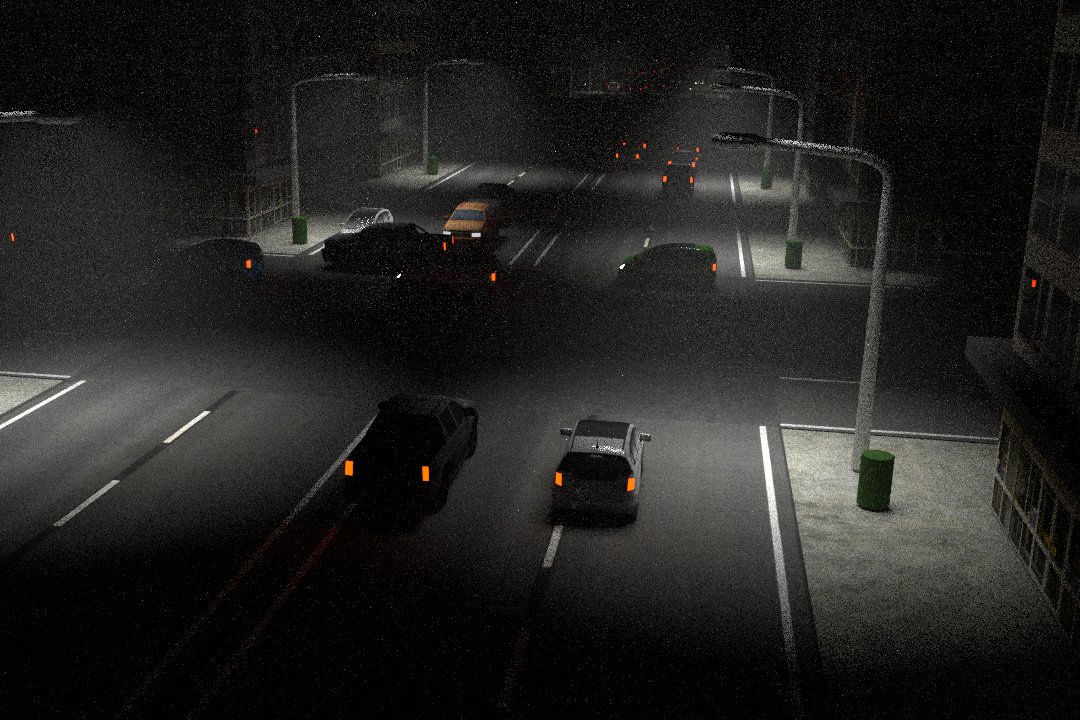}
\caption{\small night scene with fog}
\label{fig_night_fog}
\end{subfigure}
\caption{\small Domain Modeling and Data Simulation}
\label{fig:my_label}
\end{figure*}

\subsubsection{Object Shape and Texture}
\begin{figure}
\centering
\includegraphics[width=8cm, height=6.5cm]{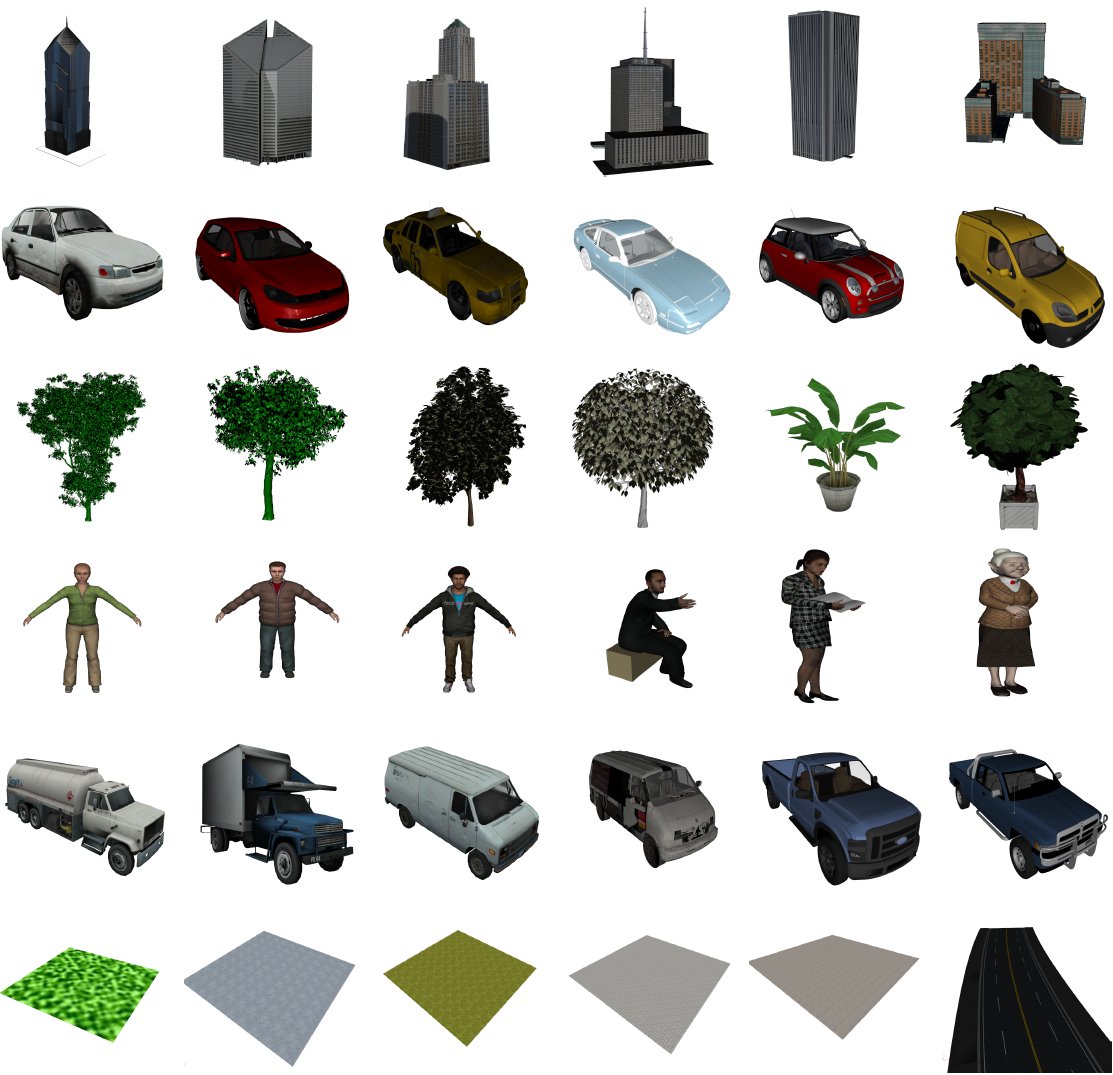}
\caption{\small Sample 3D object models - Buildings, Cars, Trees, Humans, Trucks, Ground, Road models}
\label{fig_3dmodels}
\end{figure}

 The intrinsic object shape models are randomly selected from the 3D object repositories and resized to fit in the sampled 3D bounding boxes. We have collected a rich set of 3D models for each object
category (buildings, grounds, pedestrians, and vehicles etc) from the web\footnote{3dwarehouse.sketchup.com, 3dmodelfree.com, quality3dmodels.net, tf3dm.com}. Some of the 3D object models are shown in Figure \ref{fig_3dmodels}. These 3D models are indexed according their object category.  However, the intrinsic object shape information can be modelled with the distributions \cite{Guan:Thesis:2012} such as Boltzmann machines \cite{wu20153d} or Bernoulli mixture distributions \cite{ge2009marked}.

\subsection{Photometry}
\textbf{Light models}:-
A light model (for sun light) which is parametrized by wavelength (color) and intensity, is used for the simulations in this work. A sample image rendered under sun light (mid season color) is shown in Figure \ref{fig_auto}. Another sample image rendered under white light with intensity level 4 units is shown in Figure \ref{fig_surveillance}. Spot light models are used to simulate street lights  in the scene, shown in Figure \ref{fig_night_fog}. 

\textbf{Materials/BRDFs}:-
Physics inspired BRDF (Bi-directional reflectance distribution functions) models for diffuse, specular and glassy surfaces are available with the base of the platform (Blender). We also used binary image maps (in addition to texture maps) to activate corresponding shader (diffuse or glassy). For example, see Figure \ref{fig_render_ambient} for which glassy BRDF shader is used for window surfaces on buildings. Emission BRDF is also used for some purposes in our simulations, for example, brake-lights of vehicles.

\textbf{Weather models}:- 
Light that travels through weather or any other participating medium, undergoes three kinds of phenomena: absorption, in and out scattering, and emission. Radiance of a particle that is scattered into the viewing direction, increases because of light impinging on the particle due to inscattering and emission.  The spatial distribution of the scattered light is modeled by the phase functions $p(\omega_o, \omega_i)$ and different phase functions (such as Mie and Schlick etc) have been proposed and applied to simulated polluted sky, haze, clouds, and fog etc. In this work, we use Schlick phase functions \cite{jarosz2008efficient} that are parameterized by particle size, density and anisotropy . These are proven to be well approximations for theoretical functions and well suited for Monte Carlo rendering methods. Schlick phase functions for aerosols (isotropic) and haze (anisotropic) are developed \cite{jarosz2008efficient}.  To create dynamic weather conditions such as rainy and snow scenes, we used particle systems \cite{reeves1983particle} with water droplets or snowflakes as particles. Please note that our current process of dynamic weather simulations may not be physically accurate. Dynamic weather changes the surface characteristics (rain makes it wet) and appearance depends on capture time (creates blurring effect) \cite{garg2004detection}. We neglected those effects in this work. However, these effects can be modelled approximately with \textit{DynamicPaint} feature in Blender and would be considered for future work.

\textbf{Sensor models}:-
Computer graphics domain's major focus is about photo-realistic rendering methods and they rarely consider sensor models and their influences on the rendered images. Even if the simulated images are very photo-realistic, they might not be well suited for machine vision applications unless camera imperfections are considered, which include chromatic aberration, vignetting, lens distortion, sensor noise etc. 
Hence, a physics inspired sensor with noise models \cite{tsin2001statistical} has been implemented using \textit{OpenGL} pinhole camera and some of the noise and lens effects are done using post-processing processes on the \textit{framebuffer}. For this work, we used the sensor settings similar to the one discussed in \cite{narasimhan2003models}.

\subsection{Dynamics}
We manually scripted the temporal variations of the parameters of lights, weather and objects in the scene, for the validation experiments discussed in the main article. To simulate global or local illumination changes, corresponding light source's intensity levels are increased gradually over the sequence. For weather changes, particle density and anisotropy parameters are varied. Automatic path generation and obstacle avoidance for animating the camera, pedestrians and vehicles (dynamic objects) is one of our future interests.

\section{Rendering} \label{sec_rendering}
\begin{figure*}
\centering
\begin{subfigure}[t]{1.0\textwidth}
\includegraphics[width=4.2cm, height=3cm]{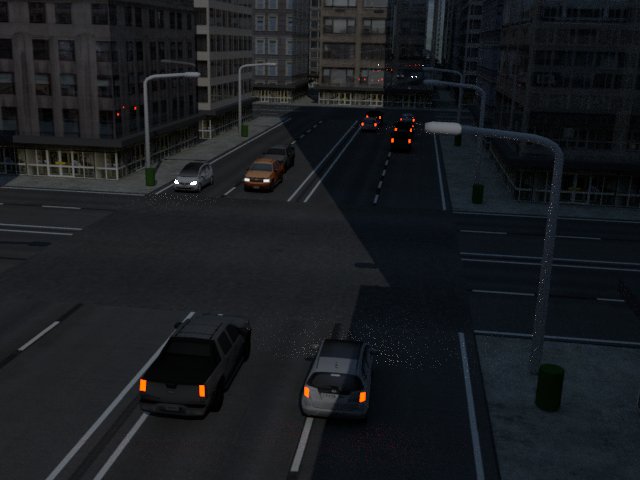}
\includegraphics[width=4.2cm, height=3cm]{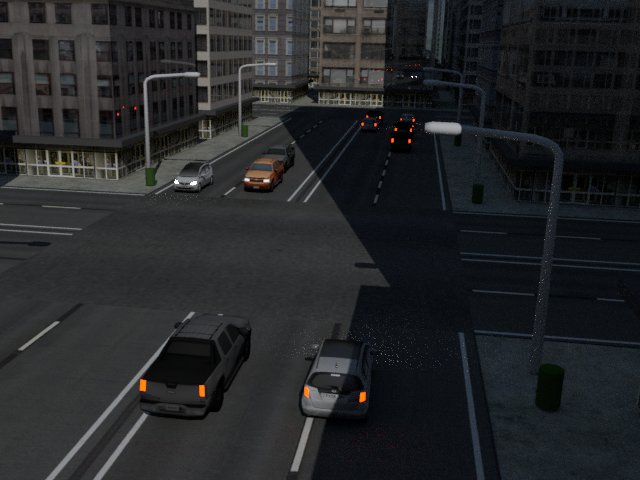}
\includegraphics[width=4.2cm, height=3cm]{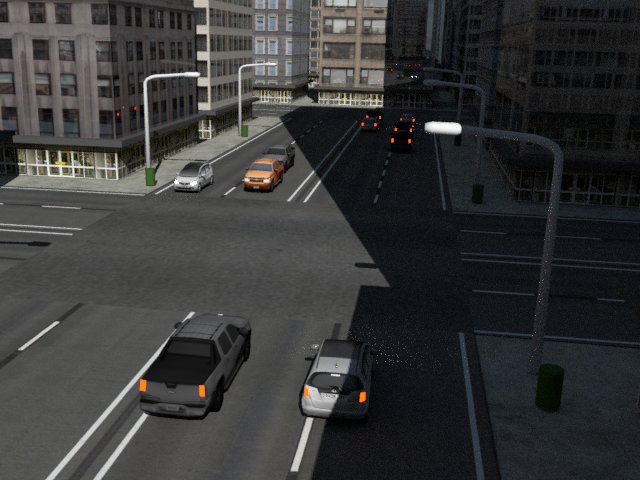}
\includegraphics[width=4.2cm, height=3cm]{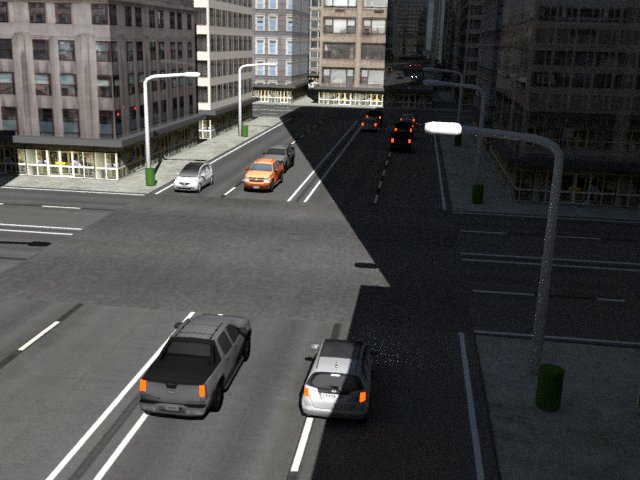}
\caption{\small Global light intensity variations}
\end{subfigure}

\begin{subfigure}[t]{1.0\textwidth}
\includegraphics[width=4.2cm, height=3cm]{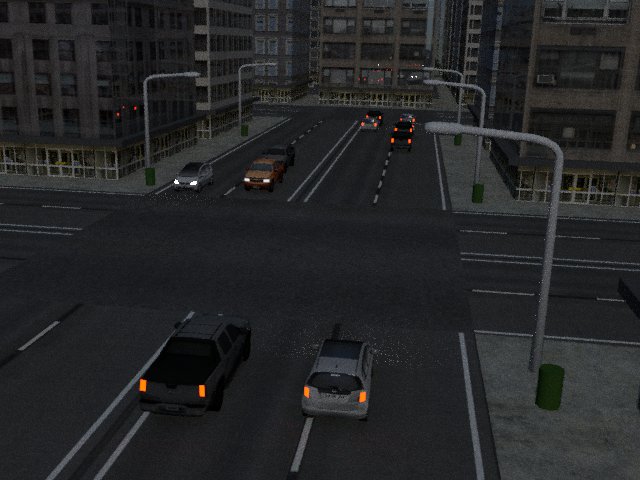}
\includegraphics[width=4.2cm, height=3cm]{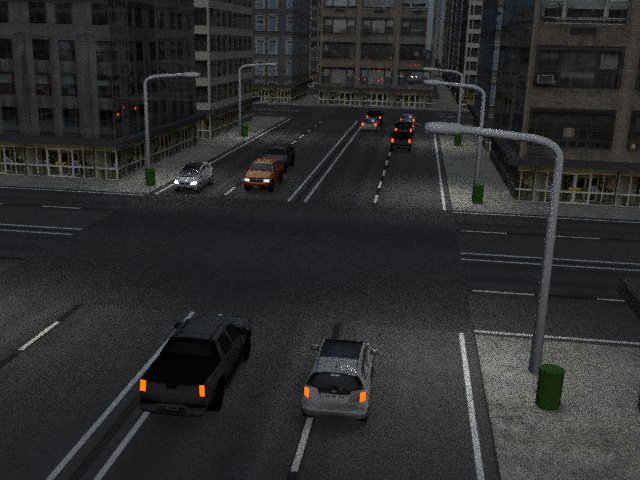}
\includegraphics[width=4.2cm, height=3cm]{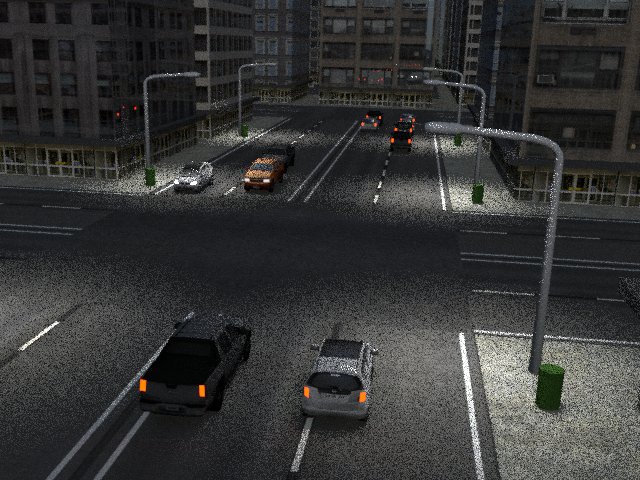}
\includegraphics[width=4.2cm, height=3cm]{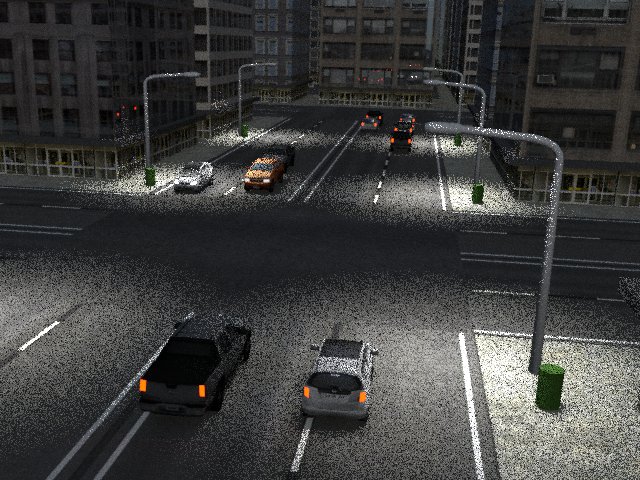}
\caption{\small Night light variations}
\end{subfigure}
\begin{subfigure}[t]{1.0\textwidth}
\includegraphics[width=4.2cm, height=3cm]{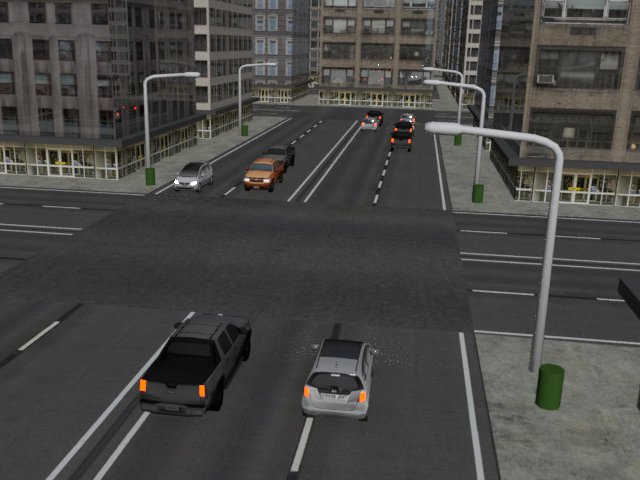}
\includegraphics[width=4.2cm, height=3cm]{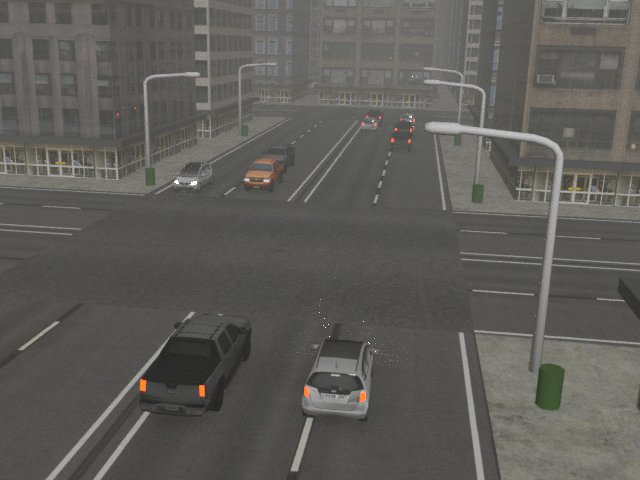}
\includegraphics[width=4.2cm, height=3cm]{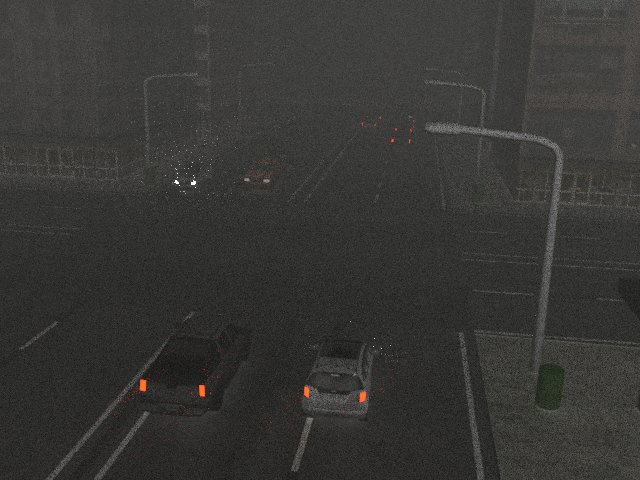}
\includegraphics[width=4.2cm, height=3cm]{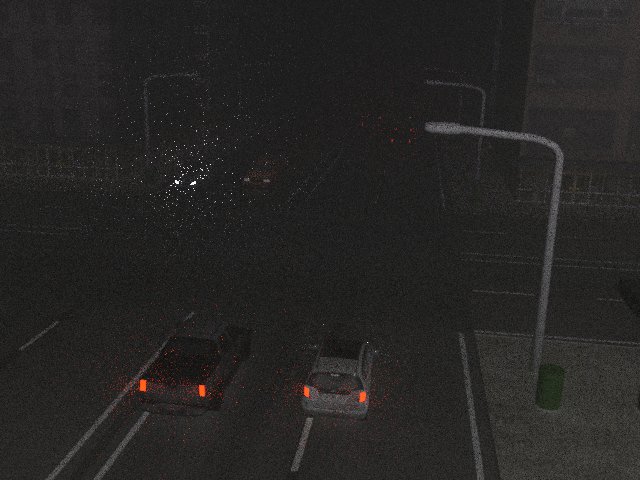}
\caption{\small Weather variations}
\label{fig_bad_weather_sim}
\end{subfigure}
\caption{\small Simulated data samples}
\label{fig_data_simulated}
\end{figure*}

\begin{figure*}
\centering
\begin{subfigure}[t]{1.0\textwidth}
\includegraphics[width=4.2cm, height=2.8cm]{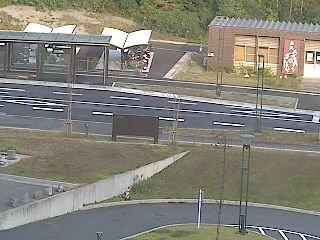}
\includegraphics[width=4.2cm, height=2.8cm]{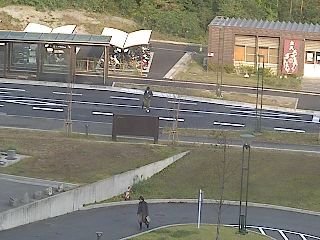}
\includegraphics[width=4.2cm, height=2.8cm]{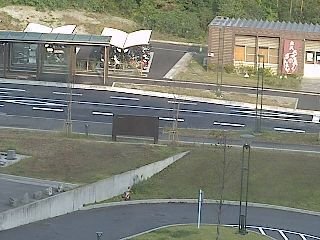}
\includegraphics[width=4.2cm, height=2.8cm]{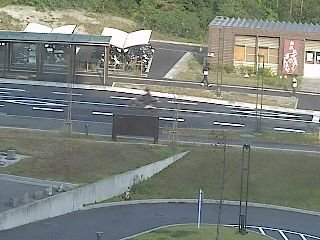}
\caption{\small Global light intensity variations, selected from \cite{shimada2014case}}
\end{subfigure}

\begin{subfigure}[t]{1.0\textwidth}
\includegraphics[width=4.2cm, height=2.8cm]{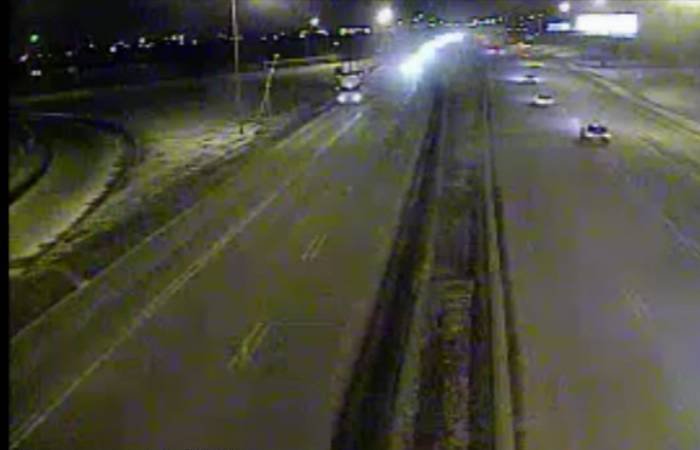}
\includegraphics[width=4.2cm, height=2.8cm]{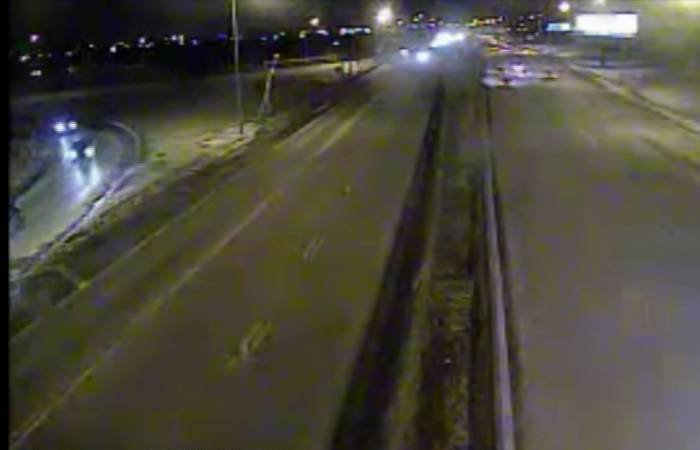}
\includegraphics[width=4.2cm, height=2.8cm]{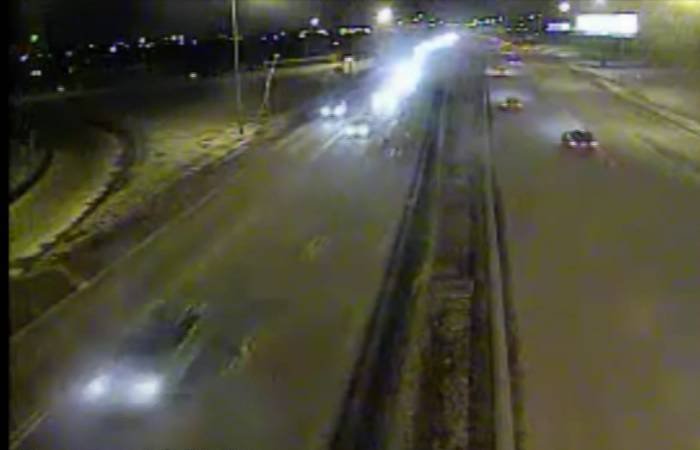}
\includegraphics[width=4.2cm, height=2.8cm]{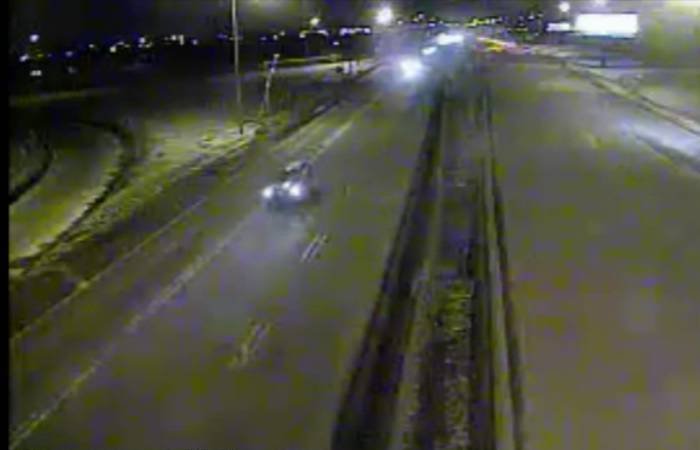}
\caption{\small Night light variations, selected from \cite{goyette2012changedetection}}
\end{subfigure}
\begin{subfigure}[t]{1.0\textwidth}
\includegraphics[width=4.2cm, height=2.8cm]{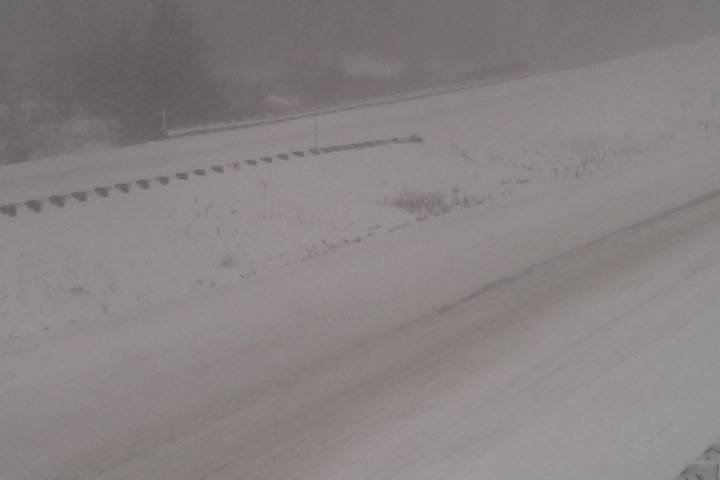}
\includegraphics[width=4.2cm, height=2.8cm]{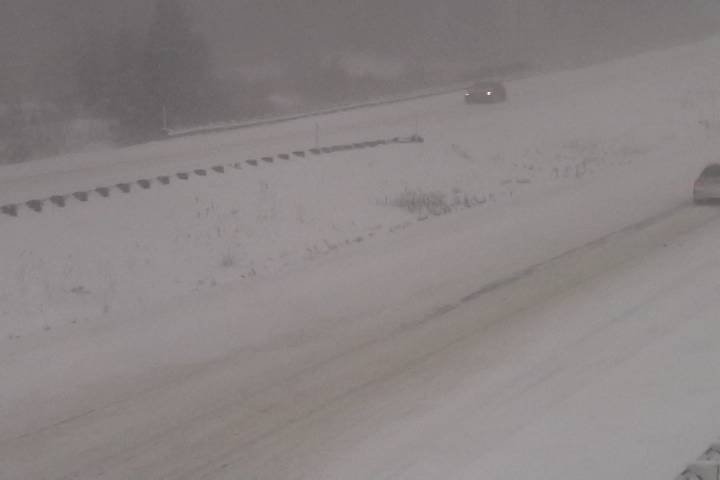}
\includegraphics[width=4.2cm, height=2.8cm]{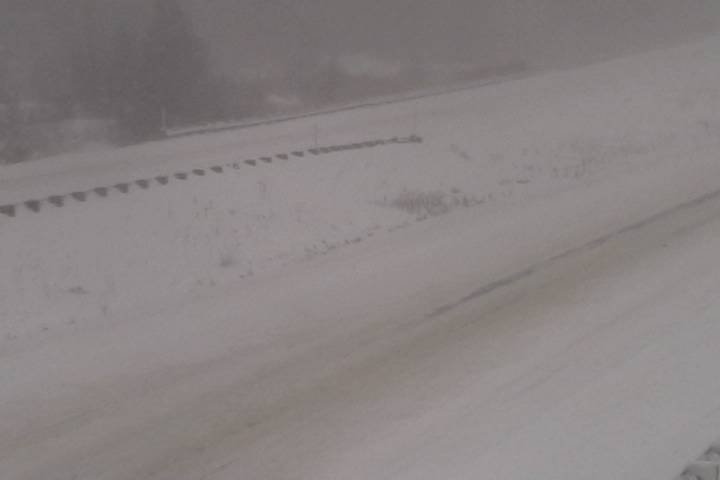}
\includegraphics[width=4.2cm, height=2.8cm]{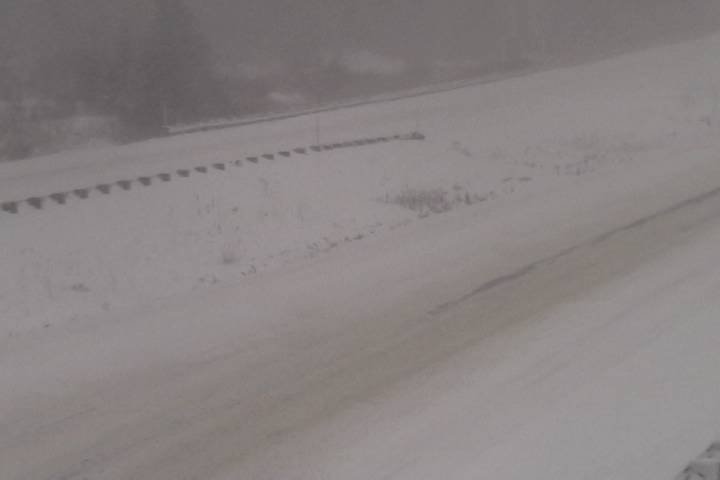}
\caption{\small Bad Weather variations, selected from \cite{goyette2012changedetection}}
\end{subfigure}
\caption{\small Real world data samples}
\label{fig_data_real}
\end{figure*}

A physics based simulation platform is discussed in \cite{Veeravasarapu:2015sv}, which facilitates sampling from domain models and rendering the images/videos along with required annotations. Rendering has been done using this platform with MC path tracing method (by allowing 200 render samples).  Some samples are shown in Figure \ref{fig_data_simulated}. 
To assess the reliability of the insights coming from the simulations, we compared these insights to that of some real world video sequences chosen from change detection benchmark datasets \cite{shimada2014case,goyette2012changedetection}, which are captured under different temporal variations, similar to the ones considered in the domain modeling (see Figure \ref{fig_data_real}). 

\section{Model Validation} \label{sec_ps}

\textbf{Piece-wise Smooth Flow Model}:
In addition to the models validated in the main article, here we consider one more hypothesis that is often used to design regularizers for motion estimators i.e, local smoothness in velocity fields. This piece-wise smooth flow (PS) constraint can either be applied solely to spatial domain or spatio-temporal domain. This is given by,
\begin{equation}
p_{PS}(u,v) = \operatornamewithlimits{\Pi}_{(i,j)\in s} \phi (|\bigtriangledown_3 u_{ij}|^2 + |\bigtriangledown_3 v_{ij}|^2)
\end{equation}

where $\bigtriangledown_3 = (\frac{\partial}{\partial x}, \frac{\partial}{\partial y}, \frac{\partial}{\partial t})$. Various functional forms (Gaussian, robust, etc.) have been assumed for the potential $\phi$. For this work, its variance is used as a measure to its behavior and established as function of context and scale. 
\begin{equation}
\sigma_{PS}^2 = f_{PS}(\theta_W, s)
\end{equation}

\begin{figure}
\includegraphics[width=8cm, height=4cm]{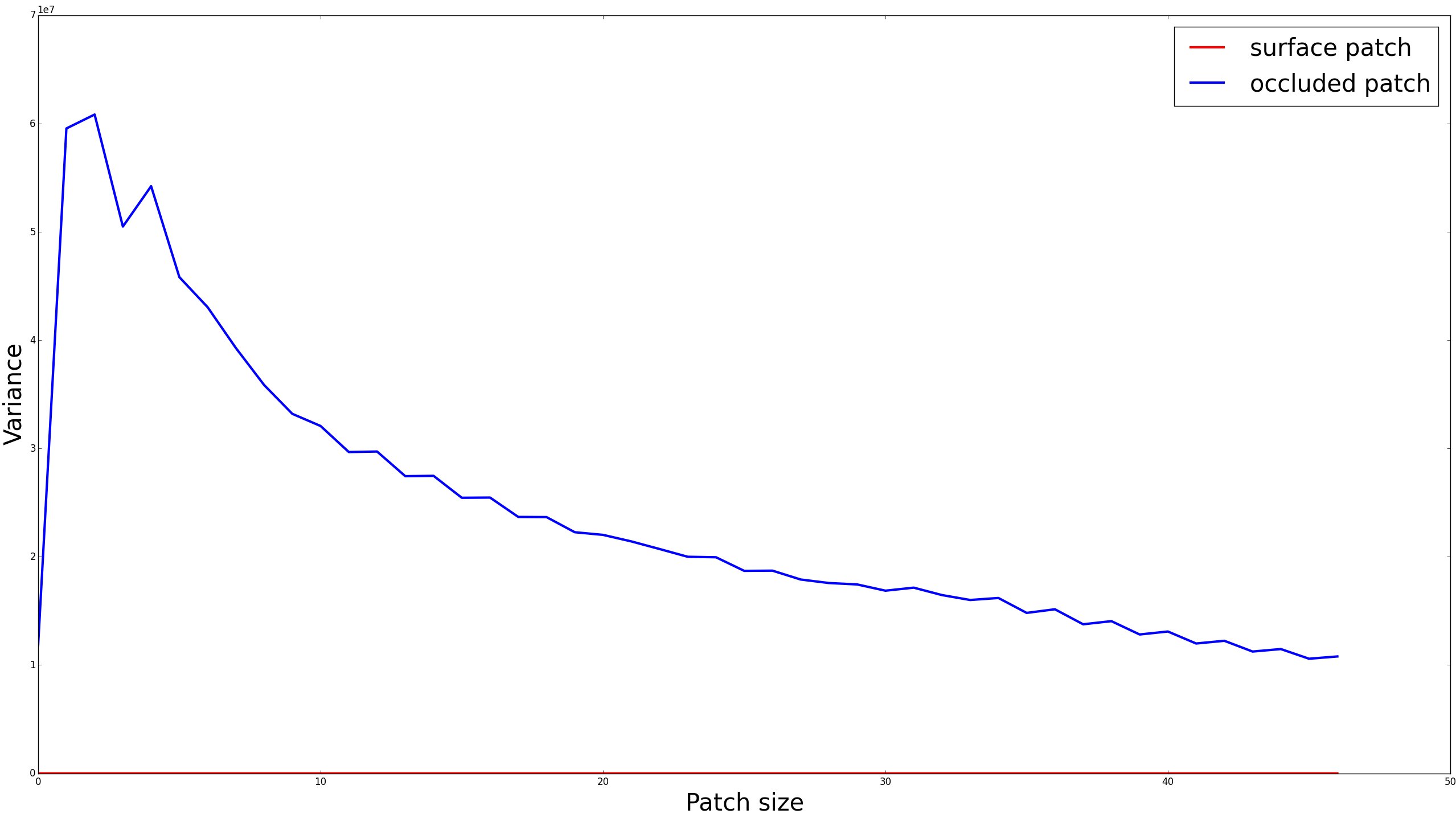}
\caption{\small Validation of PS model for patches on single object surface and occluding surfaces}
\label{fig_model_ps}
\end{figure}

We consider two types of spatial contexts (in patches): one patch contains pixels of same single object's surface and the other contains occluding surfaces (motion boundary). We manually locate these patches on the simulated images, and compute variance ($\sigma^2_{PS}$) on these patches of varying sizes (see Figure \ref{fig_model_ps}). As expected this model is valid for the patches on the same surface and violated for occlusions and motion boundaries. These qualitative insights that we got about BC, GC, and PS models (most-used models in optical flow algorithms) from our experiments match with statements and experiments found in the optical flow literature \cite{sun2010secrets}. Moreover, the histograms of pixel level statistics such as intensity, gradients, flow vectors are compared between a synthetic optical flow dataset and lookalike real world videos \cite{butler2012naturalistic}. They reported that Kullback-Liebler divergence between these distributions are minimum. 
 
Piece-wise smoothness assumptions can be applied to many features such as flow, surface normals, depth, curvatures and surface color etc. If the selected feature representation is appearance invariant (i.e. the representation depends mainly on shape and motion), then approximations in appearance due to rendering pipeline choices will not affect the feature representation. Success in transfer learning depends on closeness of the domain models used in virtual world to the reality.  Validation experiments for the models based on appearance-dependent (BC and DS) or appearance-quasi-invariant feature descriptors (OC, GC, LBP, and HOG) are presented in the main article. As mentioned already in the main article, in our specific context of experiments, the considered models are meeting qualitative expectations on the overall performance across patch types, but when one considers the relation between spatial contexts (i.e. patch types) and model's behavior the results from virtual world data and real data have minor to moderate deviations. Qualitative inferences such as ranking the models like object shape feature models in the given context are quite similar across real and virtual worlds. However the deviations and bias in the conclusions from these models can be corrected to some extent using domain adaptation concepts.

\section{Conclusions}
We presented the details of the domain modelling and rendering processes used for the data simulation for validation experiments provided in the main article. Piecewise-Smooth Flow model and its validation in virtual world is also discussed.

\end{document}